\documentclass[conference]{IEEEtran}
\IEEEoverridecommandlockouts
\usepackage{cite}
\usepackage{amsmath,amssymb,amsfonts}
\usepackage{algorithmic}
\usepackage{graphicx}
\usepackage{textcomp}
\usepackage{xcolor}

\usepackage{booktabs} 
\usepackage{graphicx}
\usepackage{subfigure}
\usepackage{amsthm}
\usepackage{amsmath}
\usepackage{multirow}
\usepackage{enumerate}
\usepackage{amsfonts}
\usepackage{color}
\usepackage{url}
\usepackage{tabularx}
\usepackage{diagbox}
\usepackage{longtable}
\usepackage{rotating}
\usepackage{slashbox}
\usepackage[justification=centering]{caption}
\usepackage[ruled,linesnumbered]{algorithm2e}
\usepackage{bm}
\usepackage{amssymb}
\usepackage{algorithmic}
\usepackage{url}

\usepackage{color}

\def\BibTeX{{\rm B\kern-.05em{\sc i\kern-.025em b}\kern-.08em
    T\kern-.1667em\lower.7ex\hbox{E}\kern-.125emX}}
\begin{document}

\title{Efficient Reinforced Feature Selection via Early Stopping Traverse Strategy
}

\author{\IEEEauthorblockN{Kunpeng Liu\IEEEauthorrefmark{1}, Pengfei Wang\IEEEauthorrefmark{2}, Dongjie Wang\IEEEauthorrefmark{1}, Wan Du\IEEEauthorrefmark{3},  Dapeng Oliver Wu\IEEEauthorrefmark{4}, Yanjie Fu\IEEEauthorrefmark{1}}
\IEEEauthorblockA{\textit{\IEEEauthorrefmark{1} University of Central Florida, Orlando, United States} \\
\textit{\IEEEauthorrefmark{2} DAMO Academy, Alibaba Group, China} \\
\textit{\IEEEauthorrefmark{3} University of California, Merced, United States} \\
\textit{\IEEEauthorrefmark{4} University of Florida, Gainesville, United States} \\
\IEEEauthorrefmark{1}\{kunpengliu, wangdongjie\}@knights.ucf.edu, yanjie.fu@ucf.edu \\ 
\IEEEauthorrefmark{2}\{wpf2106\}@gmail.com,
\IEEEauthorrefmark{3}\{wdu3\}@ucmerced.edu,
\IEEEauthorrefmark{4}\{dpwu\}@ufl.edu \\
}

\IEEEcompsocitemizethanks{
	\IEEEcompsocthanksitem \textbullet ~ Pengfei Wang is a co-first author.
	\IEEEcompsocthanksitem \textbullet ~ Yanjie Fu is the corresponding author.
	}

}

\maketitle

\begin{abstract}
In this paper, we propose a single-agent Monte Carlo based reinforced feature selection (MCRFS) method, as well as two efficiency improvement strategies, i.e., early stopping (ES) strategy and reward-level interactive (RI) strategy. Feature selection is one of the most important technologies in data prepossessing, aiming to find the optimal feature subset for a given downstream machine learning task. Enormous research has been done to improve its effectiveness and efficiency. Recently, the multi-agent reinforced feature selection (MARFS) has achieved great success in improving the performance of feature selection. However, MARFS suffers from the heavy burden of computational cost, which greatly limits its application in real-world scenarios. In this paper, we propose an efficient reinforcement feature selection method, which uses one agent to traverse the whole feature set, and decides to select or not select each feature one by one. Specifically, we first develop one behavior policy and use it to traverse the feature set and generate training data. And then, we evaluate the target policy based on the training data and improve the target policy by Bellman equation. Besides, we conduct the importance sampling in an incremental way, and propose an early stopping strategy to improve the training efficiency by the removal of skew data. In the early stopping strategy, the behavior policy stops traversing with a probability inversely proportional to the importance sampling weight. In addition, we propose a reward-level interactive strategy to improve the training efficiency via reward-level external advice. Finally, we design extensive experiments on real-world data to demonstrate the superiority of the proposed method.
\end{abstract}
\section{Introduction}
In general data mining and machine learning pipelines, before proceeding with machine learning tasks, people need to preprocess the data first. Preprocessing technologies include data cleaning, data transformation and feature engineering. As one of the most important feature engineering technique, feature selection aims to select the optimal feature subset from the original feature set for the downstream task.
Traditional feature selection methods can be categorized into three families: (i) filter methods, in which features are ranked by a specific score (e.g., univariate feature selection \cite{yang1997comparative,forman2003extensive}, correlation based feature selection \cite{hall1999feature,yu2003feature}); 
(ii) wrapper methods, in which optimal feature subset is identified by a search strategy that collaborates with predictive tasks (e.g., evolutionary algorithms \cite{yang1998feature,kim2000feature}, branch and bound algorithms \cite{narendra1977branch,kohavi1997wrappers}); (iii) embedded methods, in which feature selection is part of the optimization objective of predictive tasks (e.g., LASSO \cite{tibshirani1996regression}, decision tree \cite{sugumaran2007feature}). However, these studies have shown not just strengths but also some limitations. 
For example, filter methods ignore the feature dependencies and interactions between feature selection and predictors.
Wrapper methods have to directly search a very large feature space of  $2^N$ feature subspace candidates, where $N$ is the number of features. 
Embedded methods are subject to the strong structured assumptions of predictive models, i.e., in LASSO, the non-zero weighted features are considered to be important.

\begin{figure}[htbp]
\centering
\subfigure[Feature subspace candidates]{
\includegraphics[width=4.1cm]{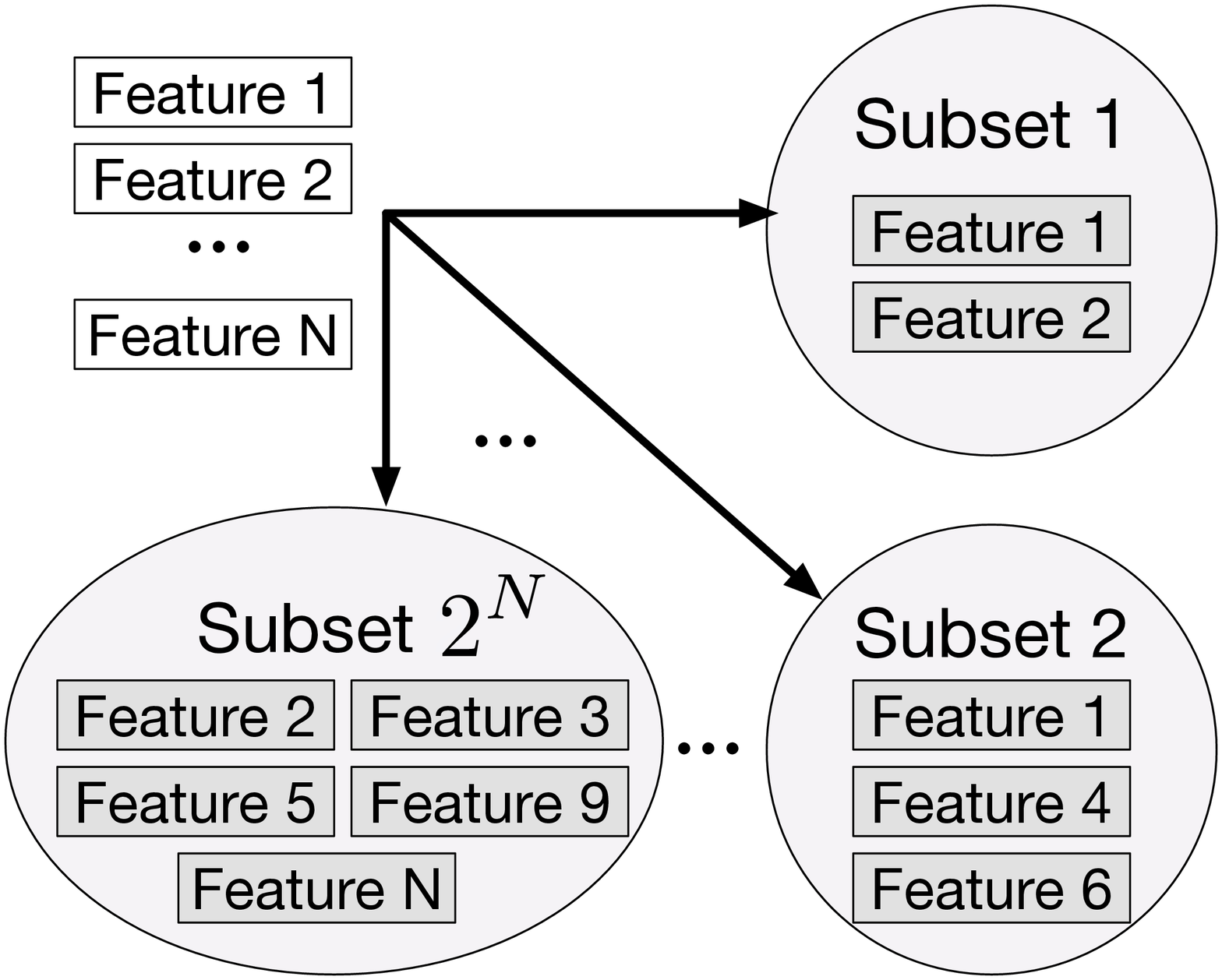}
}
\hspace{-5mm}
\subfigure[Reinforced feature selection]{
\includegraphics[width=4.1cm]{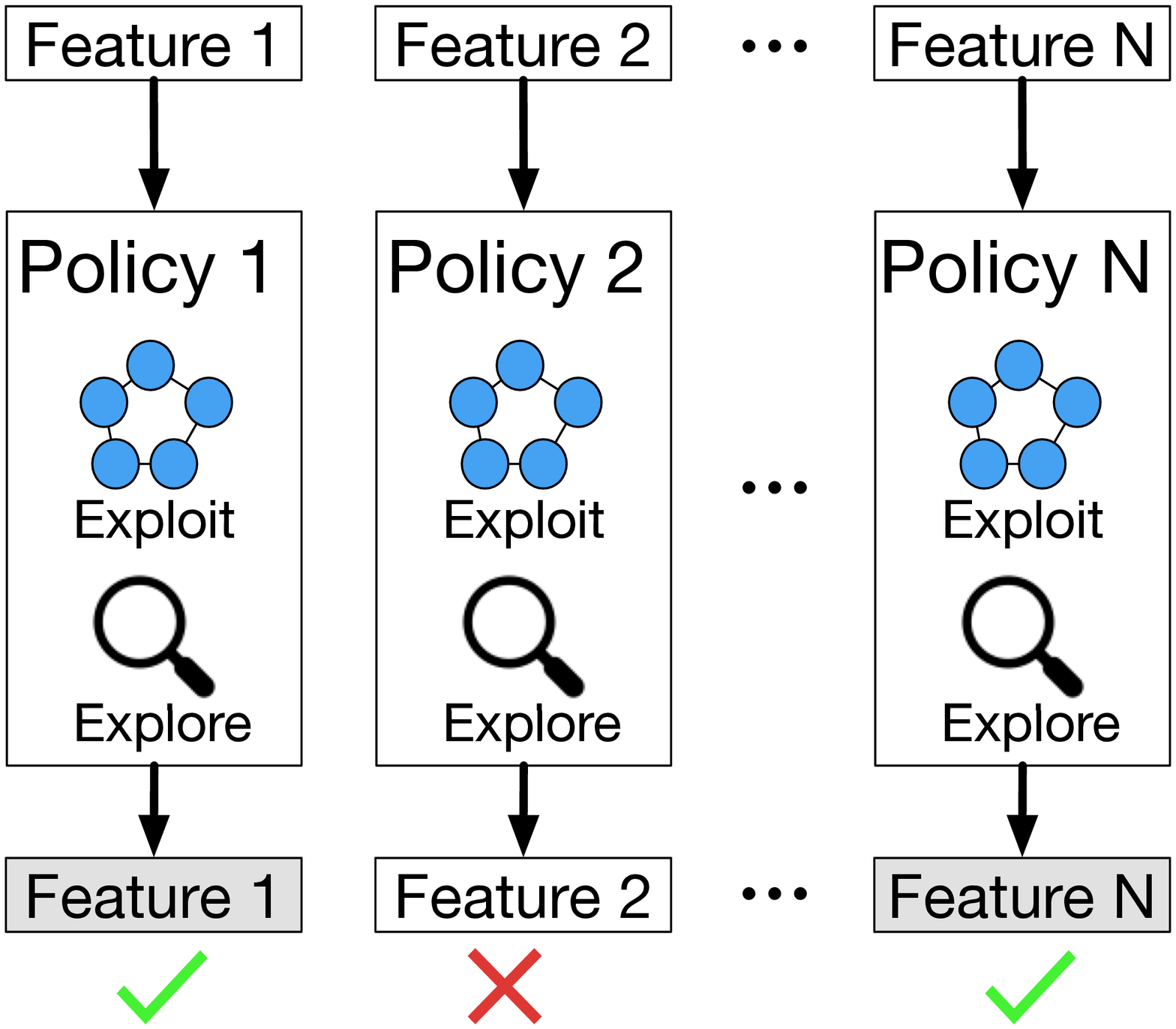}
}
\caption{Reinforced feature selection explores the feature subspace by assigning each feature one agent, and the agent's policy decides the selection of its corresponding feature.}
\label{fig_FSC_RFS}
\end{figure}

Recently, reinforcement learning has been incorporated with feature selection and produces an emerging feature selection method, called reinforced feature selection \cite{liu2019automating,fan2020autofs}. In the reinforced feature selection, there are multiple agents to control the selection of features, one agent for one feature. All agents cooperate to generate the optimal feature set. It has been proved to be 
superior to traditional feature selection methods due to its powerful global search ability. However, each agent adapts a neural network as its policy. Since the agent number equals the feature number ($N$ agents for $N$ features), when the feature set is extremely large, we need to train a large number of neural networks, which is computationally high and not applicable for large-scale datasets. Our research question is: Can we design a more practical and efficient method to address the feature selection problem while preserving the effectiveness of reinforced fearure selection? To answer this questions, there are three challenges.


The first challenge is to reformulate the feature selection problem with smaller number of agents. Intuitively, we can define the action of the agent as the selected feature subset. For a given feature set, we input it to the agent's policy and the agent can directly output the optimal subset. However, the feature subset space is as large as $2^N$, where $N$ is the feature number. When the dataset is large, the action space is too large for the agent to explore directly.
To tackle this problem, we design a traverse strategy, where one single agent visit each feature one by one to decide its selection (to select or deselect). After traversing all the feature set, we can obtain the selected feature subset.
We adapt the off-policy Monte Carlo method to our framework. In the implementation, we design two policies, i.e., one behavior policy and one target policy. The behavior policy is to generate the training data and the target policy is to generate the final feature subset. In each training iteration, we use the behavior policy to traverse the feature set, and generates one training episode. The training episode consists of a series of training samples, each of which contains the state, the action and the reward. Similar with \cite{liu2019automating}, we regard the selected feature subset as the environment, and its representation as the state. The action 1/0 denotes selection/deselection, and the reward is composed of predictive accuracy, feature subset relevance and feature subset redundancy. Using the training episode, we evaluate the target policy by calculating its Q value with importance sampling, and improve it by the Bellman equation. After more and more iterations, the target policy becomes better and better. After the training is done, we use the target policy to traverse the feature set and can derive the optimal feature subset. Besides, the behavior policy is supposed to cover the target policy as much as possible so as to generate more high-quality training data, and should introduce randomness to enable exploration \cite{sutton2018reinforcement}. We design an $\epsilon$-greedy behavior policy, to better balance the coverage and the diversity. 

The second challenge is to improve the training efficiency of the proposed traverse strategy. In this paper, we improve the efficiency from two aspects. One improvement is to conduct the importance sampling in an incremental way, which saves repeated calculations between samples. In the off-policy Monte Carlo method, since the reward comes from the behavior policy, when we use it to evaluate the target policy, we need to multiply it by an importance sampling weight. We decompose the sampling weight into an incremental format, where the calculation of the sampling weight can directly use the result of previous calculations. The other improvement is to propose an early stopping criteria to assure the quality of training samples as well as stopping the meaningless traverse by behavior policy. In Monte Carlo method, if the behavior policy is too far away from the target policy, the samples from the behavior policy are considered harmful to the evaluation of the target policy. As the traverse method is continuous and the importance sampling weight calculation depends on the previous result, once the sample at time $t$ is skew, the following samples are skew. We propose a stopping criteria based on the importance sampling weight, and re-calculate a more appropriate weight to make the samples from the behavior policy more close to the target policy.

The third challenge is how to improve the training efficiency by external advice. In classic interactive reinforcement learning, the only source of reward is from the environment, and the advisor does have access to the reward function. However, in many cases, the advisor can not give direct advice on action, but can evaluate the state-action pair. In this paper, we define a utility function $\mathcal{U}$ which can evaluate state-action pair and provide feedback to the agent just like the environment reward does. 
When integrating the advisor utility $\mathcal{U}$ with the environment reward $\mathcal{R}$ to a more guiding reward $\mathcal{R}^\prime$, we should not change the optimal policy, namely the optimal policy guided by $\mathcal{R}^\prime$ should be identical to the optimal policy guided by $\mathcal{R}$. In this paper, we propose a state-based reward integration strategy, which leads to a more inspiring integrated reward as well as preserving the optimal policy.

To summarize, the contributions of this paper are: (1) We reformulate the reinforced feature selection into a single-agent framework by proposing a traverse strategy; (2) We design an off-policy Monte Carlo method to implement the proposed framework; (3) We propose an early stopping criteria to improve the training efficiency. (4) We propose a reward-level interactive strategy to improve the training efficiency. (5) We design extensive experiments to reveal the superiority of the proposed method.

\begin{table}[htp!]
\centering
\caption{Commonly Used Notations.}
\renewcommand\tabcolsep{10.0pt}
\begin{tabular}{p{0.08\textwidth}p{0.31\textwidth}}

\hline
Notations & Definition \\
\hline
$s_t$, $a_t$ & state at time $t$ and action at time $t$ \\
$s^i$, $a^j$ & the $i$-th state and the $j$-th action\\
$\mathcal{S}$ & state space defined as $\{s^i| i<inf\}$\\
$\mathcal{A}$ & action space defined as $\{a^j| j \in [1, N]\}$\\
$\gamma$ & discount factor in range [0,1]\\
$\mathcal{P}(s_t, a_t, s_{t+1})$ & transition probability\\
$\pi(s) / \pi^*(s)$ & policy/optimal policy \\
$\mathcal{M}$ & Markov decision process (MDP) defined as $\{\mathcal{S},\mathcal{A},\mathcal{R}, \gamma, \mathcal{P}\}$\\
$\mathcal{U}$ & utility function from advisor's perspective\\
$\mathcal{F}$ & feature space (set) defined as $\{\mathbf{f}^k|k\in[0,M]\}$\\
\hline
\end{tabular}
\label{method0}
\end{table}

\section{Preliminaries}\label{Preliminaries}
We first introduce some preliminary knowledge about the Markov decision process(MDP) and the Monte Carlo method to solve MDP, then we give a brief description of multi-agent reinforced feature selection.

\subsection{Markov Decision Process}


Markov decision process (MDP) is defined by a tuple $\mathbf{M} = \{\mathcal{S},\mathcal{A},\mathcal{R}, \gamma, \mathcal{P}\}$, where state space $\mathcal{S}$ is finite, action space $\mathcal{A}$ is pre-defined, reward function $\mathcal{R}: \mathcal{S} \times \mathcal{A} \rightarrow \mathbb{R} $ is a mapping function from state-action pair to a scalar, $\gamma \in [0,1]$ is a discount factor and $\mathcal{P}: \mathcal{S} \times \mathcal{A} \times \mathcal{S} \rightarrow \mathbb{R}$ is the transition probability from state-action pair to the next state. In this paper, we study the most popular case when the environment is deterministic and thus $\mathcal{P} \equiv 1$.
We use superscripts to discriminate different episode, and use subscripts to denote the time step inside the episode, e.g., $s^i_t$, $a^i_t$ denote the state and action at time $t$ in the $i$-th episode.

\subsection{Monte Carlo for Solveing MDP}
Monte Carlo method can take samples from the MDP to evaluate and improve its policy. Specifically, at the $i$-th iteration, with a behavior (sampling) policy $b^i$, we can derive an episode $\mathbf{x}^i = \{x^i_1, x^i_2, \dots, x^i_t, \dots x^i_N\}$, where $x^i_t = (s^i_t, a^i_t, r^i_t)$ is a sample consisting state, action and reward. With the episode, we can evaluate the Q value $Q_{\pi^i}(s^i_t, a^i_t)$ over our policy (detailed in Section \ref{MCRFS}), and improve it by Bellman optimality: 
\begin{equation}\label{Bellman_Eq}
    \pi^{i+1}(s) = argmax_a Q_{\pi^i}(s, a)
\end{equation}
With the evaluation-improvement process going on, the policy $\pi$ becomes better and better, and can finally converge to the optimal policy. The general process is:
\begin{equation}\label{improve}
    \pi^0 \stackrel{E}{\rightarrow} Q_{\pi^0} \stackrel{I}{\rightarrow} \pi^1 \stackrel{E}{\rightarrow} Q_{\pi^1} \stackrel{I}{\rightarrow} \dots \pi^M \stackrel{I}{\rightarrow} Q_{\pi^M}
\end{equation}
where $\stackrel{E}{\rightarrow}$ denotes the policy evaluation and $\stackrel{I}{\rightarrow}$ denotes the policy improvement. After $M$ iterations, we can achieve an optimal policy.
As Equation \ref{improve} shows, the policy evaluation and improvement need many iterations, and each iteration needs one episode $\mathbf{x}$ ($\mathbf{x}^i$ for the $i$-th iteration) consisting $N$ samples.

\subsection{Multi-Agent Reinforced Feature Selection}
\begin{figure}[t]
\centering
\includegraphics[width=6cm]{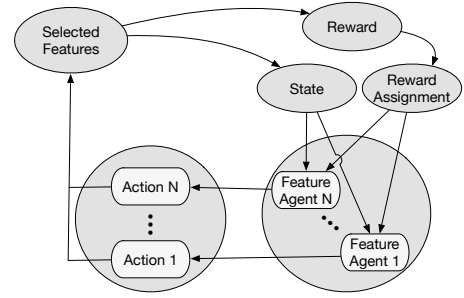}
\caption{Multi-agent Reinforced feature selection. Each feature is controlled by one feature agent.}
\label{fig_MARL}
\end{figure}

Feature selection aims to find an optimal feature subset $\mathcal{F}^\prime$ from the original feature set $\mathcal{F}$ for a downstream machine learning task $\mathcal{M}$.
Recently, the emerging multi-agent reinforced feature selection (MARFS) method \cite{liu2019automating} formulates the feature selection problem into a multi-agent reinforcement learning task, in order to automate the selection process. 
As Figure \ref{fig_MARL} shows, in the MARFS method, each feature is assigned to a feature agent, and the action of feature agent decides to select/deselect its corresponding feature. It should be noted that the agents simultaneously select features, meaning that there is only one time step inside an iteration, and thus we omit the subscript here. At the $i$-th iteration, all agents cooperate to select a feature subset $\mathcal{F}^i$. The next state $s^{i+1}$ is derived by the representation of selected feature subset $\mathcal{F}^i$:

\begin{equation}
    s^{i+1} = represent(\mathcal{F}^i)
\end{equation}
where $\mathcal{F}^i$ is the selected feature subset at time $t$. $represent$ is a representation learning algorithm which converts the dynamically changing $\mathcal{F}^i_t$ into a fixed-length state vector $s^{i+1}$. The $represent$ method can be meta descriptive statistics, autoencoder based deep representation and dynamic-graph based GCN in \cite{liu2019automating}.
The reward $r^i$ is an evaluation of the selected feature subset $\mathcal{F}^i$:
\begin{equation}
    r^i = eval(\mathcal{F}^i)
\end{equation}
where $eval$ is evaluations of $\mathcal{F}^i$, which can be a supervised metric with the machine learning task $\mathcal{F}$ taking $\mathcal{F}^i$ as input, unsupervised metrics of $\mathcal{F}^i$, or the combination of supervised and unsupervised metrics in \cite{liu2019automating}. The reward is assigned to each of the feature agent to train their policies. With more and more steps' exploration and exploitation, the policies become more and more smart, and consequently they can find better and better feature subsets. 

\section{Proposed Method}
In this section, we first propose a single-agent Monte Carlo based reinforced feature selection method. And then, we propose an episode filtering method to improve the sampling efficiency of the Monte Carlo method. In addition, we apply the episode filtering Mote Carlo method to the reinforced feature selection scenario. Finally, we design a reward shaping strategy to improve the training efficiency.

\begin{figure*}[!th]
\centering
\includegraphics[width=18cm]{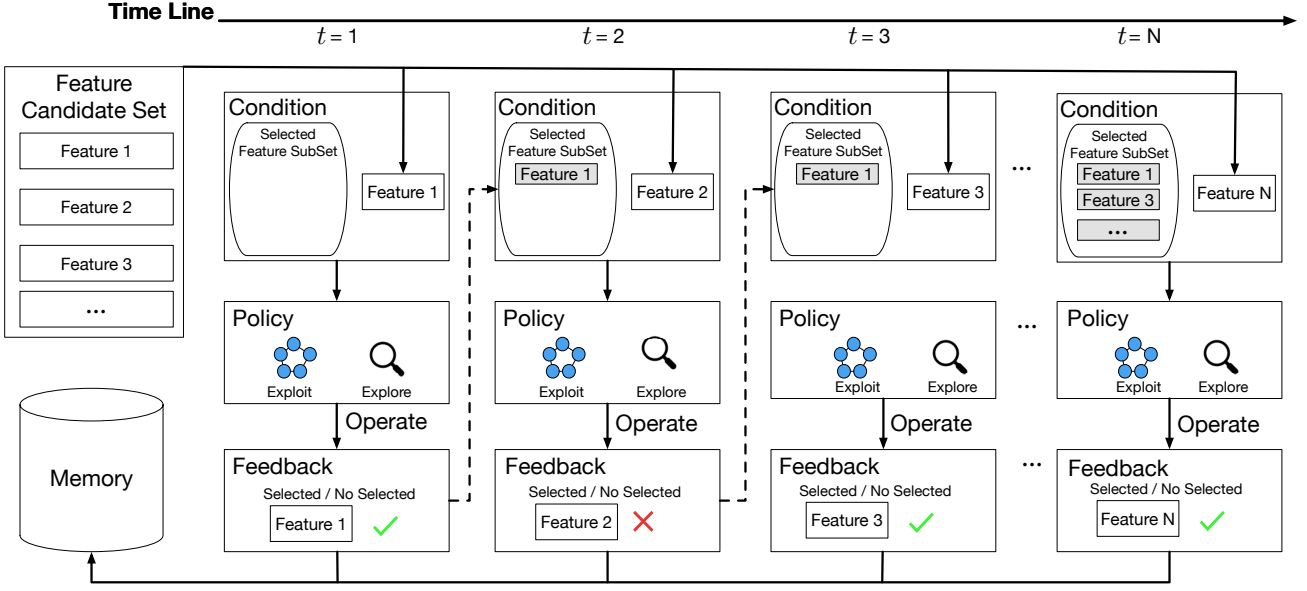}
\vspace{-2.5mm}
\caption{Single-agent Reinforced feature selection with traverse strategy. At each step, the agent transverses \\ features one by one to decide their selection. The traverse data are stored in the memory to form a training episode. }
\vspace{-3mm}
\label{fig_SRFS}
\end{figure*}

\subsection{Monte Carlo Based Reinforced Feature Selection}\label{MCRFS}
The MARFS method has proved its effectiveness, however, the multi-agent strategy greatly increases the computational burden and hardware cost. Here, we propose a single-agent traverse strategy and use Mote Carlo method as the reinforcement learning algorithm.

\subsubsection{Traverse strategy}
As Figure \ref{fig_SRFS} shows, rather than using $N$ agents to select their corresponding features in the multi-agent strategy, we design one agent to traverse all features one at a time.  

In the $i$-th episode, beginning from time $t=1$, the behavior policy $b^i$ firstly decides the selection decision (select or not select) for feature 1, and then, at time $t=2$, $b^i$ decides the selection decision for feature 2. With time going on, the features are traversed one by one, and the selected features forms a selected feature subset $\mathcal{F}^i_t$. Meanwhile, this process also generates an episode $\mathbf{x}^i_N = \{x^i_1, x^i_2, \dots, x^i_t, \dots x^i_N\}$, where $x^i_t = (s^i_t, a^i_t, r^i_t)$ is a tuple of state, action and reward. The action $a^i_t = 1/0$ is the selection/deselection decision of the $t$-th feature, the next state $s^i_{t+1}$ is derived by $represent(\mathcal{F}^i_t)$ and the reward $r^i_t$ is derived by $eval(\mathcal{F}^i_t)$.

\subsubsection{Monte Carlo Method for Reinforced Feature Selection}
With the episode generated by the behavior policy $b^i$, we can evaluate our target policy $\pi^i$ and improve $\pi^i$. Both the behavior policy $b^i$ and the target policy $\pi^i$ provide the probability of taking action $a$ given a specific state $s$.

Specifically, We generate an episode $\mathbf{x}^i_N$ by $b^i$. Then, we calculate the accumulated reward by:
\begin{equation}
    G^i(s^i_t, a^i_t) = \sum_{j=0}^t \gamma^{(t-j)} \cdot r^i_j
\end{equation}
where $0 \leq \gamma \leq 1$ is a discount factor. 

As the state space is extremely large, we use a neural network $Q(s, a)$ to approximate $G(s, a)$. 

The target policy $\pi$ is different from the behavior policy $b$, and the reward comes from samples derived from policy $b$, therefore the accumulated reward of $\pi$ should be calculated by multiplying an importance sampling weight:

\begin{equation}\label{overall_weight}
\rho_t^i = \frac{\Pi_{j=0}^{t}\pi^i(a^i_j|s^i_j)}{\Pi_{j=0}^{t}b^i(a^i_j|s^i_j)}
\end{equation}

The $Q_{\pi^i}(s, a)$ can be optimized by minimizing the loss:

\begin{equation}\label{loss_pi}
    \mathcal{L}_{\pi^i} = || Q_{\pi^i}(s^i_t, a^i_t) - \rho_t^i*G^i(s^i_t, a^i_t)||^2 
\end{equation}

The probability of taking action $a$ for state $s$ under policy $\pi$ in the next iteration can be calculated by:

\begin{equation}\label{pro_pi}
    \pi^{i+1}\{a|s\} = \frac{exp(Q_{\pi^i}\{a, s\})}{exp(Q_{\pi^i}\{a = 0, s\}) + exp(Q_{\pi^i}\{a = 1, s\})}
\end{equation}

We develop an $\epsilon$-greedy policy of $b$ based on the Q value from $\pi$:
\begin{equation}\label{pro_b}
    b^{i+1}\{a|s\}=
    \begin{cases}
    1-\epsilon&  a = argmax_a Q_{\pi^i}(s,a);\\
    0 & otherwise;
    \end{cases}
\end{equation}

Algorithm \ref{MCRFS} shows the process of Monte Carlo based feature selection (MCRFS) with traverse strategy. 


\begin{algorithm}[t]
\caption{Monte Carlo Based Reinforced Feature Selection with Traverse Strategy}
\label{MCRFS}
\LinesNumbered 

\KwIn{Feature set $\mathcal{F} = \{f_1, f_2, ..., f_N \}$, downstream machine learning task $\mathcal{T}$.}
\KwOut{Optimal feature subset $\mathcal{F}^\prime$.}
Initialize the behavior policy $b^1$, target policy $\pi^1$, exploration number $M$, $\mathcal{F}^\prime = \Phi$. 

\For{$i=1$ to $M$}
{
    Initialize state $s^i_1$.\\
    \For{$t=1$ to $N$}
    {
        Derive action $a^i_t$ with behavior policy $b^i(s^i_t)$.\\

        Perform $a^i_t$, getting selected feature subset $\mathcal{F}^i_t$.\\
        
        Obtain the next state $s^i_{t+1}$ by $represent(\mathcal{F}^i_t)$ and reward $r^i_t$ by $eval(\mathcal{F}^i_t)$.\\
        }
        
        Update target policy $\pi^{i+1}$ by Equation \ref{pro_pi} and behavior policy $b^{i+1}$ by Equation \ref{pro_b}.\\
            
        \If{$eval(\mathcal{F}^i_N)>eval(\mathcal{F}^\prime)$}{$\mathcal{F}^\prime = \mathcal{F}^i_N$.}
}
Return $\mathcal{F}^\prime$.
\end{algorithm}

\subsection{Early Stopping Monte Carlo Based Reinforced Feature Selection} \label{EFMC}

In many cases, the feature set size $N$ can be very large, meaning that there can be a large number of samples in one episode $\mathbf{x}^i_N$. The problem is, if the sample at time $T$ is bad (the Chi-squared distance between $b^i(s^i_t)$ and $\pi^i(s^i_t)$ is large), all the subsequent samples (from $T$ to $N$) in the episode are skew \cite{maceachern1999sequential}. The skew samples not only are a waste time time to generate, but also do harm to the policy evaluation, therefore we need to find some way to stop the sampling when the episode becomes skew.

\subsubsection{Incremental Importance Sampling}

Rather than calculating the importance sampling weight for each sample directly by Equation \ref{overall_weight}, we here decompose it into an incremental format. Specifically, in the $i$-th iteration, we define the weight increment:

\begin{equation}
    w_t^i = \frac{\pi^i(a^i_t|s^i_t)}{b^i(a^i_t|s^i_t)}
\end{equation}

and the importance sampling weight can be calculated by:
\begin{equation}
    \rho_t^i = \rho_{t-1}^i \cdot w_t^i
\end{equation}
Thus, at each time, we just need to calculate a simple increment to update the weight. 
\subsubsection{Early Stopping Monte Carlo Method for Reinforced Feature Selection}
We first propose the stopping criteria, and then propose a decision history based traversing strategy to enhance diversity.

\noindent \textbf{\textsl{Early stopping criteria}}. We stop the traverse by probability:
\begin{equation}\label{stop_pro}
    p_t^i = max(0, 1 -\rho_t^i/v)
\end{equation}
where $0 \leq v \leq 1$ is the stopping threshold. 

And for the acquired episode, we recalculate the importance sampling weight for each sample by:

\begin{equation}\label{pvm}
    w_t^i = p_v^i\cdot \rho_t^i/p_t^i
\end{equation}
where the $p_v$ can be calculated by:
\begin{equation}
    p_v^i = \int max (0, 1 -\rho_t^i/v) b^i(\mathbf{s}_t) \; d\mathbf{s}_t
\end{equation}

As $p_v^i$ is identical for all samples in the $i$-th episode regardless of $t$, the calculation of Equation \ref{pvm} does almost no increase to the computation.


\noindent \textbf{\textsl{Decision history based traversing strategy}}. In the $i$-th iteration, the stopping criteria stops the traverse at time $t$, and the features after $t$ are not traversed. With more and more traverses, the front features (e.g., $f_1$ and $f_2$) are always selected/deselected by the agent, while the backside features (e.g., $f_N$ and $f_{N-1}$) get very few opportunity to be decided. To tackle this problem, we record the decision times we made on each feature, and re-rank their orders to diversify the decision process in the next traverse episode. For example, in the past $5$ episodes, if the decision times of feature set \{$f_1$, $f_2$, $f_3$\} are \{5,2,4\}, then in the 6-th episode, the traverse order is $f_2 \rightarrow f_3 \rightarrow f_1$.


Algorithm \ref{MCRFS_ES} shows the process of Monte Carlo based feature selection (MCRFS) with early stopping traverse strategy. specifically, we implement the early stopping Monte Carlo based reinforced feature selection method as follows:
\begin{itemize}
    \item[1.] Use a random behavior policy $b^0$ to traverse the feature set. Stop the traverse with the probability in Equation \ref{stop_pro} and get an episode $\mathbf{x}_{N_0}^0$. 
    \item [2.] Evaluate the policy $\pi^0$ to get the Q value $Q^0$ by minimizing Equation \ref{loss_pi}, and derive the updated policy $\pi^1$ and $b^1$ from Equation \ref{pro_pi} and \ref{pro_b} respectively.
    \item [3.] Update the record of traverse times for each feature. Re-rank feature order. The smaller times one feature was traversed, the more forward order it should get. 
    \item[4.] Use the updated policy $\pi^1$ and $b^1$ to traverse the re-ranked feature set for the next $M$ steps. Derive the policy $\pi^M$ and $b^M$. Use $\pi^M$ to traverse the feature set without stopping criteria, and derive the final feature subset.
\end{itemize}

\begin{algorithm}[t]
\caption{Monte Carlo Based Reinforced Feature Selection with Early Stopping Traverse Strategy}
\label{MCRFS_ES}
\LinesNumbered 

\KwIn{Feature set $\mathcal{F} = \{f_1, f_2, ..., f_N \}$, downstream machine learning task $\mathcal{T}$.}
\KwOut{Optimal feature subset $\mathcal{F}^\prime$.}
Initialize the behavior policy $b^1$, target policy $\pi^1$, exploration number $M$, $\mathcal{F}^\prime = \Phi$. 

\For{$i=1$ to $M$}
{
    Initialize state $s^i_1$.\\
    Rank features with their decision history. \\
    \For{$t=1$ to $N$}
    {
        Derive action $a^i_t$ with behavior policy $b^i(s^i_t)$.\\

        Perform $a^i_t$, getting selected feature subset $\mathcal{F}^i_t$.\\
        
        Obtain the next state $s^i_{t+1}$ by $represent(\mathcal{F}^i_t)$ and reward $r^i_t$ by $eval(\mathcal{F}^i_t)$.\\
        
        Break the loop with probability $p_t^i$ derived from Equation \ref{stop_pro};
        }

        Update target policy $\pi^{i+1}$ by Equation \ref{pro_pi} and behavior policy $b^{i+1}$ by Equation \ref{pro_b}.\\
            
        \If{$eval(\mathcal{F}^i_N)>eval(\mathcal{F}^\prime)$}{$\mathcal{F}^\prime = \mathcal{F}^i_N$.}
}
Return $\mathcal{F}^\prime$.
\end{algorithm}

\subsection{Interactive Reinforcement Learning}

As all the steps in this section belong to the same iteration, we omit the superscript $i$ in each denotation for simplicity.

Reinforcement learning is proposed to develop the optimal policy $\pi_\mathcal{M}^*(s) = argmax_{a} Q_{\mathcal{M}}^*(s,a)$ for an MDP $\mathcal{M}$.
The optimal $Q$-value can be updated by Bellman equation \cite{bellman1957dynamic}:

\begin{equation}\label{Bellman_Equ}
    Q_{\mathcal{M}}^*(s_t,a_t) = \mathcal{R}(s_t, a_t) + \gamma * max_{a_{t+1}}Q_{\mathcal{M}}^*(s_{t+1},a_{t+1})
\end{equation}

\begin{figure}[h]
\centering
\includegraphics[width=8.5cm]{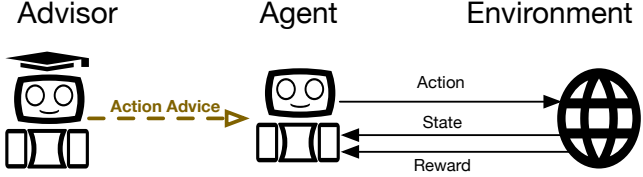}

\caption{Classic interactive reinforcement learning. The advisor gives the agent advice at the action level.}

\label{fig_act_IRL}
\end{figure}

Interactive reinforcement learning (IRL) is proposed to accelerate the learning process of reinforcement learning (RL) by providing external action advice to the RL agent \cite{suay2011effect}. As Figure \ref{fig_act_IRL} shows, for selected advising states, the action of RL agent is decided by the advisor's action advice instead of its own policy. The algorithm to select advising states varies with the problem setting. Typical algorithms for selecting advising states include early advising, importance advising, mistake advising and predictive advising \cite{torrey2013teaching}. To better evaluate the utility of the state-action pair $(s_t, a_t)$, we define a utility function $\mathcal{U}(s_t, a_t)$. The utility function can give a feedback of how the action benefits from the state from the advisor's point of view. 

\noindent \textbf{\textsl{Reward-Level Interactive Reinforcement Learning}}.
In reinforcement learning (RL), we aim to obtain the optimal policy for the MDP $\mathcal{M} = \{\mathcal{S},\mathcal{A},\mathcal{R}, \gamma, \mathcal{P}\}$. However, in IRL, when we change the reward function $\mathcal{R}$ to a more inspiring reward function $\mathcal{R}^\prime$, the original MDP $\mathcal{M}$ is changed to a new MDP $\mathcal{M}^\prime = \{\mathcal{S},\mathcal{A},\mathcal{R}^\prime, \gamma, \mathcal{P}\}$. Without careful design , the optimal policy derived from $\mathcal{M}^\prime$ would be different from the optimal policy for $\mathcal{M}$. 
Here we give a universal form of reward advice without limitation on the form of utility function $\mathcal{U}(s,a)$: 
\begin{equation}\label{reward_advice}
    \mathcal{R}^\prime (a_t,s_t) = \mathcal{R}(a_t,s_t) + c * (\gamma * \mathcal{U}(s_{t+1}) -  \mathcal{U}(s_t))
\end{equation}
where $\mathcal{U}(s_t) = E_{a_t}[\mathcal{U}(a_t, s_t)]$, $c$ is the weight to balance the proportion of the utility function.

We prove that the optimal policies of $\mathcal{M}$ and $\mathcal{M}^\prime$ are identical when the reward advice is  Equation \ref{reward_advice}:

We firstly subtract  $c * \mathcal{U}(s_t)$ from both sides of Equation \ref{Bellman_Equ}, and we have:
\begin{equation}
    \begin{split}
        Q_{\mathcal{M}}^*(s_t,a_t) &- c * \mathcal{U}(s_t)= \mathcal{R}(s_t, a_t) \\
        &+\gamma * max_{a_{t+1}}Q_{\mathcal{M}}^*(s_{t+1},a_{t+1}) - c * \mathcal{U}(s_t)
    \end{split}
\end{equation}

\begin{algorithm}[t]
\caption{Reward-Level Interactive Reinforcement Learning}
\label{alg_reward_level_IRL}
\LinesNumbered 
Initialize replay memory $\mathcal{D}$; Initialize the $Q$-value function with random weights; Initialize the advising state number $N_a$, stop time $T$;

\For{$t=1$ to $T$}
{

    $a_t=
    \begin{cases}
    \text{random action}& \text{with probability} \epsilon;\\
    max_{a_t}Q(s_t, a_t)& \text{with probability} 1-\epsilon;
    \end{cases}$
    
    Perform $a_t$, obtaining reward $\mathcal{R}(a_t,s_t)$ and next state $s_{t+1}$;

    $\mathcal{R}^\prime (s_t,a_t)=
    \begin{cases}
    \mathcal{R}(s_t,a_t)& t> N_a;\\
    \mathcal{R}(s_t,a_t) + c * (\gamma * \mathcal{U}(s_{t+1}) -  \mathcal{U}(s_t))& t \leq N_a;
    \end{cases}$
    
    Store transition $(s_t, a_t, \mathcal{R}^\prime (s_t,a_t), s_{t+1})$ in $\mathcal{D}$;
    
    Randomly sample mini-batch of data from $\mathcal{D}$;
    
     Update $Q(s,a)$ with the sampled data;
}
\end{algorithm}

We add and subtract $c* \gamma * \mathcal{U}(s_{t+1})$ on the right side:
\begin{equation}\label{subtract_two_sides}
    \begin{split}
        &Q_{\mathcal{M}}^*(s_t,a_t) - c * \mathcal{U}(s_t)= \mathcal{R}(s_t, a_t)\\
        &+\gamma * max_{a_{t+1}}Q_{\mathcal{M}}^*(s_{t+1},a_{t+1}) - c * \mathcal{U}(s_t)\\
        &+ c* \gamma * \mathcal{U}(s_{t+1}) - c* \gamma * \mathcal{U}(s_{t+1})\\
        &=\mathcal{R}(s_t, a_t) + c* \gamma * \mathcal{U}(s_{t+1})- c * \mathcal{U}(s_t) \\
        &+\gamma * max_{a_{t+1}}[Q_{\mathcal{M}}^*(s_{t+1},a_{t+1}) - c * \mathcal{U}(s_{t+1})]\\
    \end{split}
\end{equation}

We define:
\begin{equation}\label{def_of_partial}
    Q_{\mathcal{M}^\prime}^{\partial}(s_t,a_t) = Q_{\mathcal{M}}^*(s_t,a_t) - c * \mathcal{U}(s_t)
\end{equation}

Then Equation \ref{subtract_two_sides} has the new form:
\begin{equation}
    \begin{split}
        &Q_{\mathcal{M}^\prime}^{\partial}(s_t,a_t)= \mathcal{R}(s_t, a_t) + c* [\gamma * \mathcal{U}(s_{t+1})- \mathcal{U}(s_t)] \\
        &+\gamma * max_{a_{t+1}}Q_{\mathcal{M}^\prime}^{\partial}(s_{t+1},a_{t+1})\\
        & = \mathcal{R^\prime}(s_t, a_t)+\gamma * max_{a_{t+1}}Q_{\mathcal{M}^\prime}^{\partial}(s_{t+1},a_{t+1})
    \end{split}
\end{equation}
which is the Bellman equation of $Q_{\mathcal{M}^\prime}^{\partial}(s_t,a_t)$ with reward $\mathcal{R}^\prime$, meaning $Q_{\mathcal{M}^\prime}^{\partial}(s_t,a_t)$ is the optimal policy $Q$-value for MDP $\mathcal{M}^\prime$, i.e.,
\begin{equation}\label{partial_eqv_opt}
    Q_{\mathcal{M}^\prime}^*(s_t,a_t) = Q_{\mathcal{M}^\prime}^{\partial}(s_t,a_t)
\end{equation}

We combine Equation \ref{def_of_partial} and Equation \ref{partial_eqv_opt} and have:
\begin{equation}
    Q_{\mathcal{M}^\prime}^*(s_t,a_t) = Q_{\mathcal{M}}^*(s_t,a_t) - c * \mathcal{U}(s_t)
\end{equation}

Obviously,
\begin{equation}
\begin{split}
    argmax_{a_t} Q_{\mathcal{M}^\prime}^*(s_t,a_t) & =  argmax_{a_t}[Q_{\mathcal{M}}^*(s_t,a_t) - c * \mathcal{U}(s_t)]\\
    & = argmax_{a_t}Q_{\mathcal{M}}^*(s_t,a_t)
\end{split}
\end{equation}
which reveals the optimal policy of MDP $\mathcal{M}^\prime$ with reward $\mathcal{R}^\prime$ is identical to the optimal policy of MDP $\mathcal{M}$ with reward $\mathcal{R}$.

As the reward advice $\mathcal{R}^\prime$ consists of more information than the original reward $\mathcal{R}$, it can help the reinforcement learning agent explore the environment more efficiently. We give a detailed description of reward-level IRL in Algorithm \ref{alg_reward_level_IRL}. Specifically, we adapt the early advising strategy \cite{torrey2013teaching} to select the advising states, i.e., the advisor gives advice for the first $n$ states the IRL agent meets.

\begin{figure}[t]
\centering
\includegraphics[width=8.5cm]{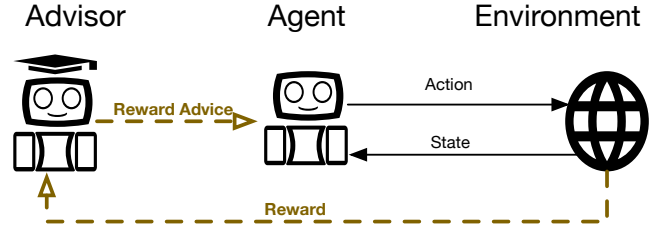}
\caption{Reward-level interactive reinforcement learning. The advisor gives advice at the reward level.}

\label{fig_overview_IRL}
\end{figure}

\subsection{Comparison with Prior Literature}
Compared with filter methods, our methods capture feature interactions; Compared with wrapper methods, our methods reduce the search space; Compared with embedded methods, our methods don't rely on strong structured assumptions; Compared with multi-agent reinforcement learning feature selection, our methods achieve parallel performance with lower computational cost.

\section{Experimental Results}

We conduct extensive experiments on real-world datasets to study: (1) the overall performance of early stopping Monte Carlo based reinforced feature selection (\textbf{ES-MCRFS}); (2) the training efficiency of the early stopping criteria; (3) the sensitivity of the threshold in the early stopping criteria; (4) the computational burden of the traverse strategy; (5) the decision history based traverse strategy; (6) the behavior policy in the ES-MCRFS. 

\subsection{Experimental Setup}

\subsubsection{Data Description}
We use four publicly available datasets on classification task to validate our methods, i.e., Forest Cover (FC) dataset \cite{forest_cover_dataset}, Spambase (Spam) dataset \cite{Dua_2019}, Insurance Company Benchmark (ICB) dataset \cite{van2000coil} and Arrhythmia (Arrhy) dataset \cite{647926}. The statistics of the datasets are in Table \ref{dataset}.

\begin{table}[h]
\begin{center} 
\caption{Statistics of datasets.}

\label{dataset}
\begin{tabular}{|p{1.0cm}|p{1.0cm}|p{1.0cm}|p{1.0cm}|p{1.2cm}|}  
\hline  
& FC &Spam &ICB & Arrhy  \\\hline  
Features& 54&   57&    86  &   274 \\  \hline  
Samples& 15120&   4601&    5000  &   452 \\  \hline  
\end{tabular}  
\end{center}
\end{table}

\begin{table*}[!th]
\centering
\caption{Overall performance.}
\label{overall_perf}
\setlength{\tabcolsep}{5mm}{
\begin{tabular}{|l|l|l|l|l|l|l|l|l|l|}
\hline
\multicolumn{2}{|l|}{\multirow{2}{*}{}} & \multicolumn{2}{c|}{FC} & \multicolumn{2}{c|}{Spam} & \multicolumn{2}{c|}{ICB} & \multicolumn{2}{c|}{Arrhy} \\ \cline{3-10} 
\multicolumn{2}{|l|}{}                  & \multicolumn{1}{c|}{Acc}        & \multicolumn{1}{c|}{F1}         & \multicolumn{1}{c|}{Acc}         & \multicolumn{1}{c|}{F1}          & \multicolumn{1}{c|}{Acc}         & \multicolumn{1}{c|}{F1}         & \multicolumn{1}{c|}{Acc}          & \multicolumn{1}{c|}{F1}          \\ \hline
\multirow{7}{*}{\begin{sideways}{{Algorithms}}\end{sideways}}   & K-Best  & 0.7904          & 0.8058          & 0.9207          & 0.8347          & 0.8783          & 0.8321          & 0.6382           & 0.6406  \\ \cline{2-10} 
                              & LASSO   & 0.8438          & 0.8493          & 0.9143          & 0.8556          & 0.8801          & 0.8507          & 0.6293           & 0.6543  \\ \cline{2-10} 
                              & GFS    & 0.8498          & 0.8350          & 0.9043          & 0.8431          & 0.9099          & 0.637          & 0.6406           & 0.6550  \\ \cline{2-10} 
                              & mRMR    & 0.8157          & 0.8241          & 0.8980          & 0.8257          & 0.8998          & 0.8423          & 0.6307           & 0.6368  \\ \cline{2-10} 
                               & RFE    & 0.8046          & 0.8175          & 0.9351          & 0.8480          & 0.9045          & 0.8502          & 0.6452           & 0.6592  \\ \cline{2-10} 
                              & MARFS    & 0.8653          & 0.8404          & 0.9219          & 0.8742          & 0.8902          & 0.8604          & 0.7238           & 0.6804  \\ \cline{2-10} 
                              & MCRFS   & 0.8688          & 0.8496          & 0.9256          & 0.8738 & 0.8956          & 0.8635          & 0.7259  & 0.7152  \\ \cline{2-10} 
                              & ES-MCRFS   & \textbf{0.8942} & \textbf{0.8750} & \textbf{0.9402} &   \textbf{0.9067}        & \textbf{0.9187} & \textbf{0.8803} & \textbf{0.7563}           & \textbf{0.7360}  \\ \hline
\end{tabular}
}
\end{table*}

\subsubsection{Evaluation Metrics}
In the experiments, we have classification as the downstream task for feature selection problem, therefore we use the two most popular evaluation metrics for classification task:

\noindent\textbf{Accuracy} is given by $Acc = \frac{TP+TN}{TP+TN+FP+FN}$, where $TP,TN, FP, FN$ are true positive, true negative, false positive and false negative for all classes. 


\noindent\textbf{F1-score} is given by $F1=\frac{2*P*R}{P+R}$, where $P=\frac{TP}{TP+FP}$ is precision and $R=\frac{TP}{TP+FN}$ is recall.

\subsubsection{Baseline Algorithms}

We compare our proposed ES-MCRFS method with the following baselines: (1) \textbf{K-Best} ranks features by unsupervised scores with the label and selects the top $k$ highest scoring features \cite{yang1997comparative}. In the experiments, we set $k$ equals to half of the number of input features.(2) \textbf{LASSO} conducts feature selection via $l1$ penalty \cite{tibshirani1996regression}. The hyper parameter in LASSO is its regularization weight $\lambda$ which is set to 0.15 in the experiments.
(3) \textbf{GFS} selects features by calculating the fitness level for each feature to generate better feature subsets via crossover and mutation \cite{leardi1996genetic}.
(4) \textbf{mRMR} ranks features by minimizing feature’s redundancy and maximizing their relevance with the label\cite{peng2005feature}.
(5) \textbf{RFE} selects features by recursively selecting smaller and smaller feature subsets \cite{granitto2006recursive}.(6)\textbf{MARFS} is a multi-agent reinforcement learning based feature selection method \cite{liu2019automating}. It uses $M$ feature agents to control the selection/deselection of the $M$ features. Besides, we also compare our method with its variant without early stopping strategy, i.e., Monte Carlo based reinforced feature selection \textbf{MCRFS}.







\subsubsection{Implementation}
In the experiments, for all deep networks, we set mini-batch size to 16 and use AdamOptimizer with a learning rate of 0.01. For all experience replays, we set memory size to 200. We set the $Q$ network in our methods as a two-layer ReLU with 64 and 8 nodes in the first and second layer. The classification algorithm we use for evaluation is a random forest with 100 decision trees. The stop time is set to 3000 steps. The state representation method in reinforced feature selection is an auto-encoder method whose encoder/decoder network is a two-layer ReLU with 128 and 32 nodes in the first and second layer. 

\subsubsection{Environmental Setup} The experiments were carried on a server with an I9-9920X 3.50GHz CPU, 128GB memory and a Ubuntu 18.04 LTS operation system.

\subsection{Overall Performance}
We compare the proposed ES-MCRFS method with baseline methods and its variant with regard to the predictive accuracy. As Table \ref{overall_perf} shows, the MCRFS, which simplify the reinforced feature selection into a single-agent formulation, achieves similar performance with the multi-agent MARFS. With the help of traverse strategy and early stopping criteria, the ES-MCRFS outperforms all the other methods.

\subsection{Sensitivity Study of Early Stopping Criteria}

\begin{figure}[htbp]
\centering
\subfigure[FC]{
\includegraphics[width=4.1cm]{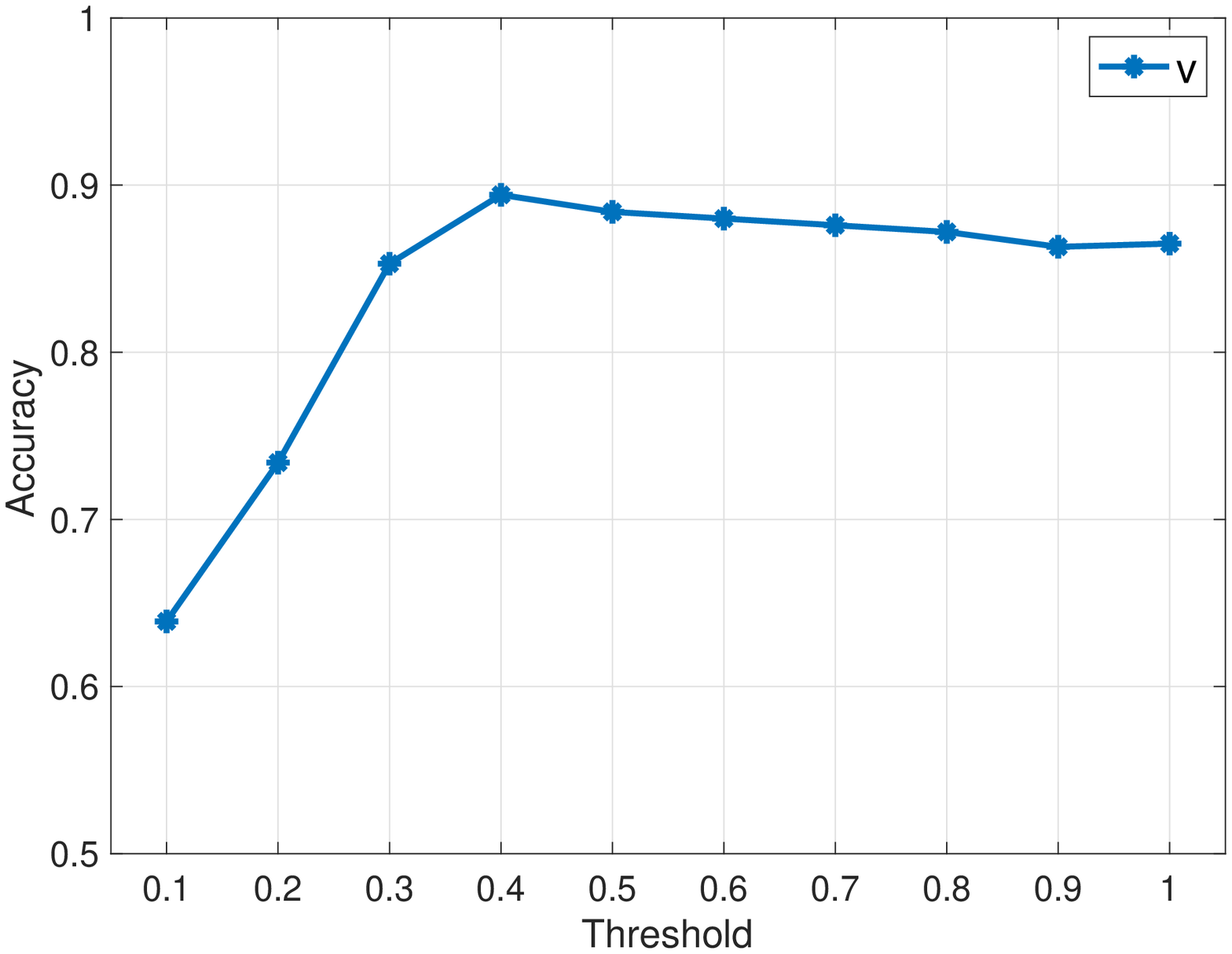}
}
\hspace{-5mm}
\subfigure[Spam]{ 
\includegraphics[width=4.1cm]{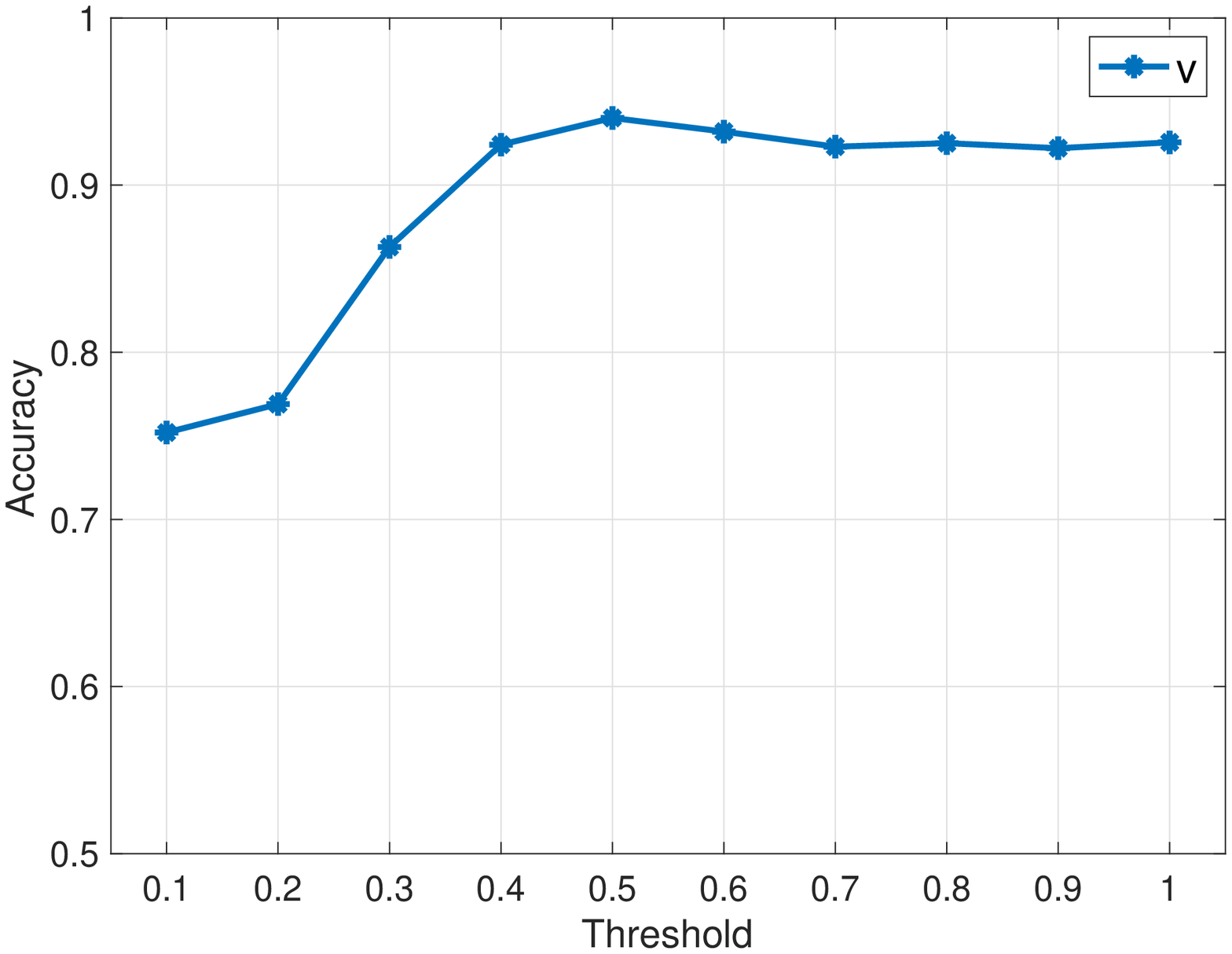}
}
\vspace{-4mm}
\subfigure[ICB]{
\includegraphics[width=4.1cm]{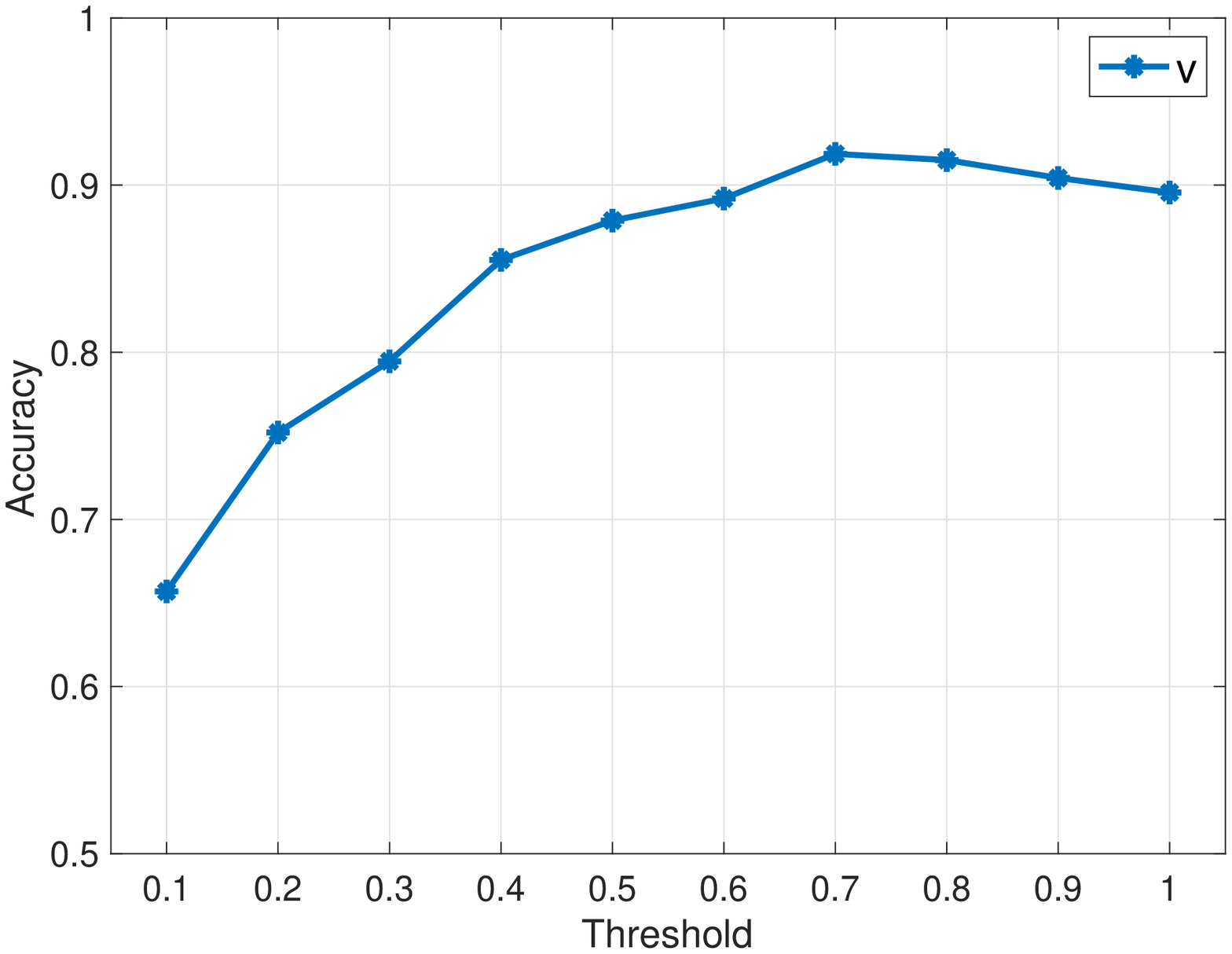}
}
\hspace{-5mm}
\subfigure[Arrhy]{ 
\includegraphics[width=4.1cm]{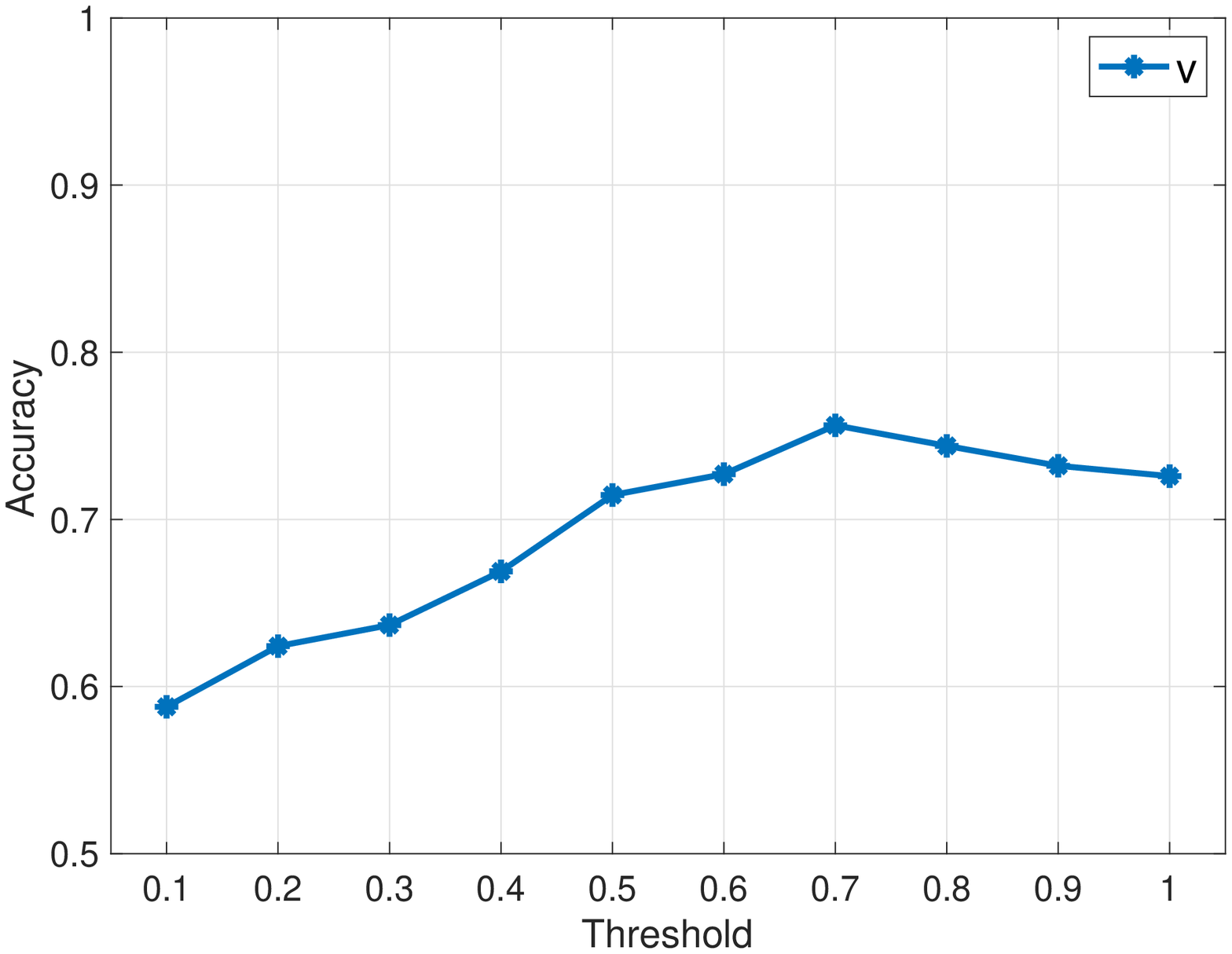}
}
\caption{Threshold sensitivity of early stopping criteria.}
\label{thresh_sens}
\end{figure}

We study the threshold sensitivity in the early stopping criteria by differing the threshold $v$ and evaluate the predictive accuracy. Figure \ref{thresh_sens} shows that the optimal threshold for the four datasets are 0.4, 0.5, 0.7, 0.7. It reveals that the early stopping criteria is sensitive to the pre-defined threshold,  and  the optimal threshold  varies on different datasets.

\subsection{Training Efficiency of Early Stopping Criteria}

\begin{figure}[p]
\centering
\subfigure[FC]{
\includegraphics[width=4.1cm]{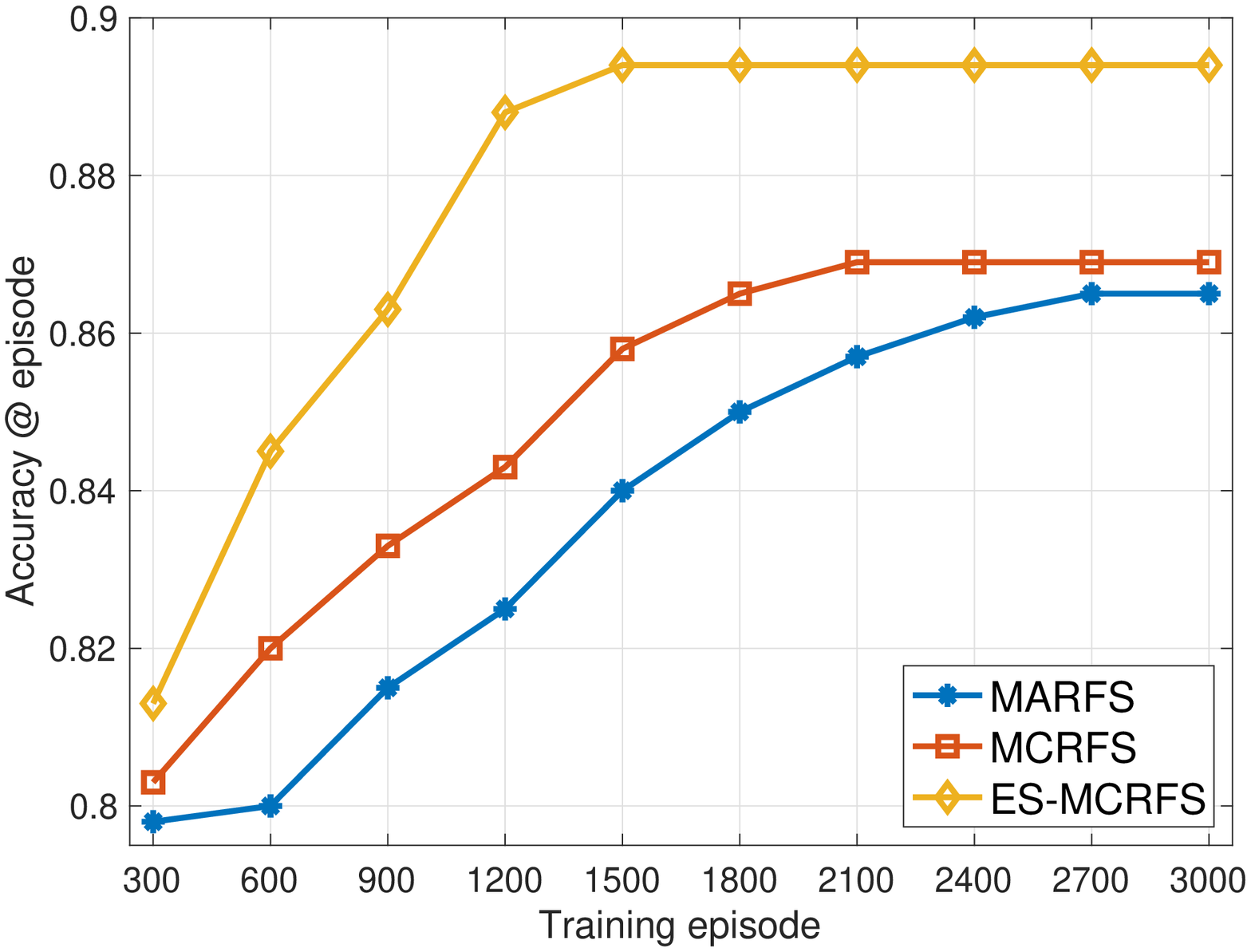}
}
\hspace{-5mm}
\subfigure[Spam]{ 
\includegraphics[width=4.1cm]{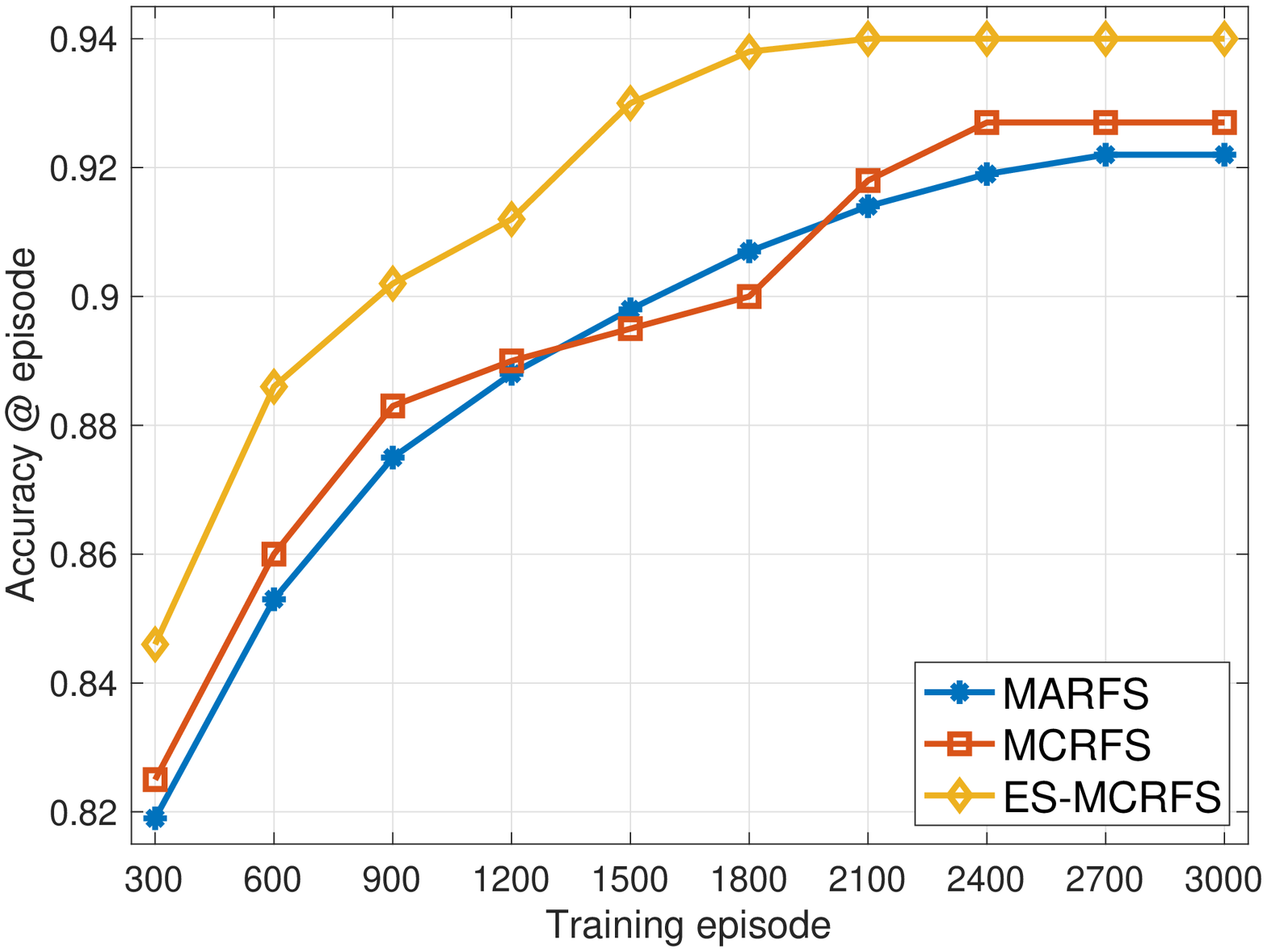}
}
\vspace{-4mm}
\subfigure[ICB]{
\includegraphics[width=4.1cm]{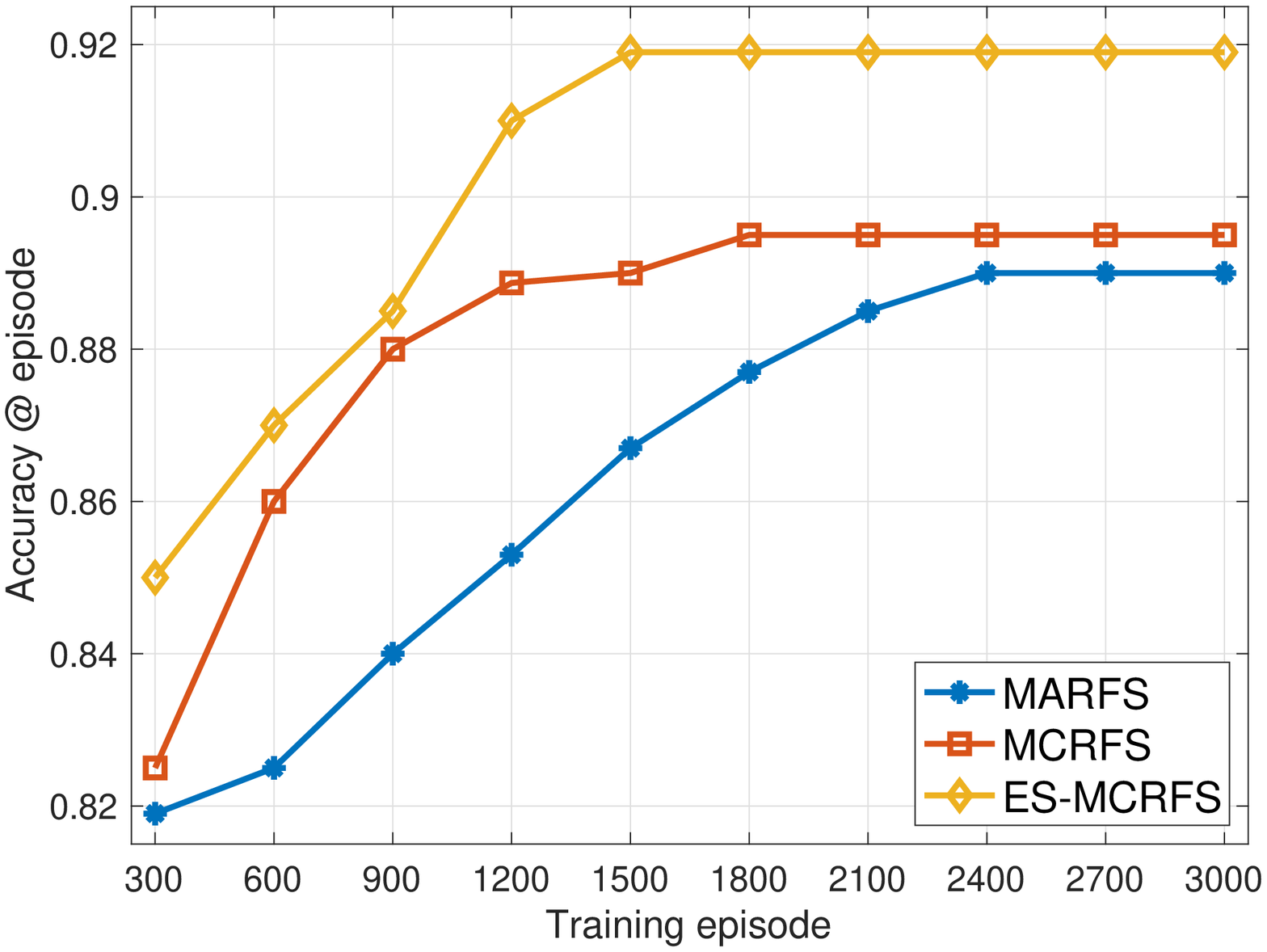}
}
\hspace{-5mm}
\subfigure[Arrhy]{ 
\includegraphics[width=4.1cm]{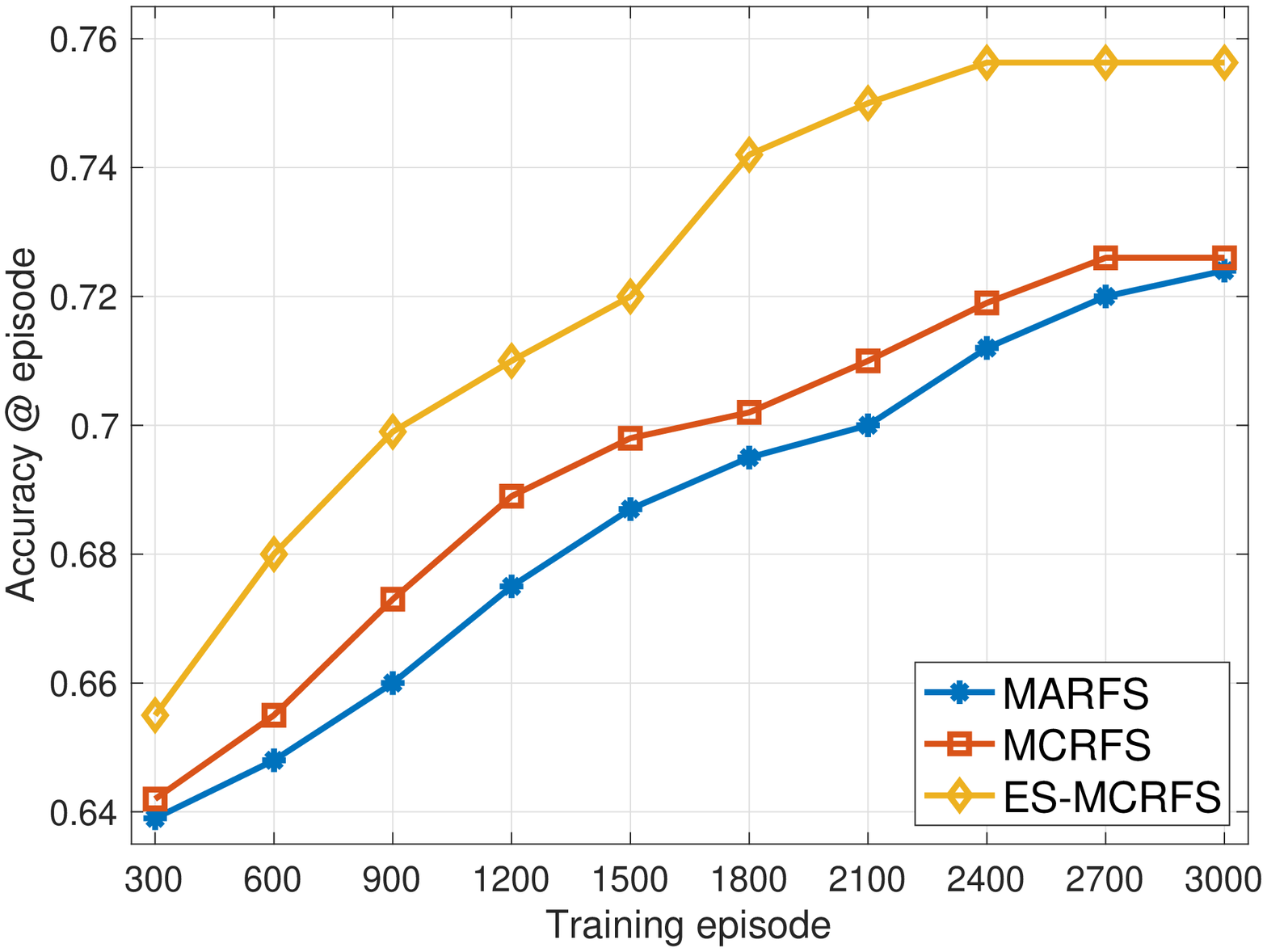}
}
\caption{Predictive accuracy on training step.}
\label{effiency_early_stop}
\end{figure}

We compare the predictive accuracy with different numbers of training episodes to study the training efficiency of the early stopping. Figure \ref{effiency_early_stop} shows that with early stopping criteria, the Monte Carlo reinforced feature selection can achieve convergence more quickly, and the predictive accuracy can be higher after convergence.

\subsection{Study of the Behavior Policy}
\begin{figure}[p]
\centering
\subfigure[FC]{
\includegraphics[width=4cm]{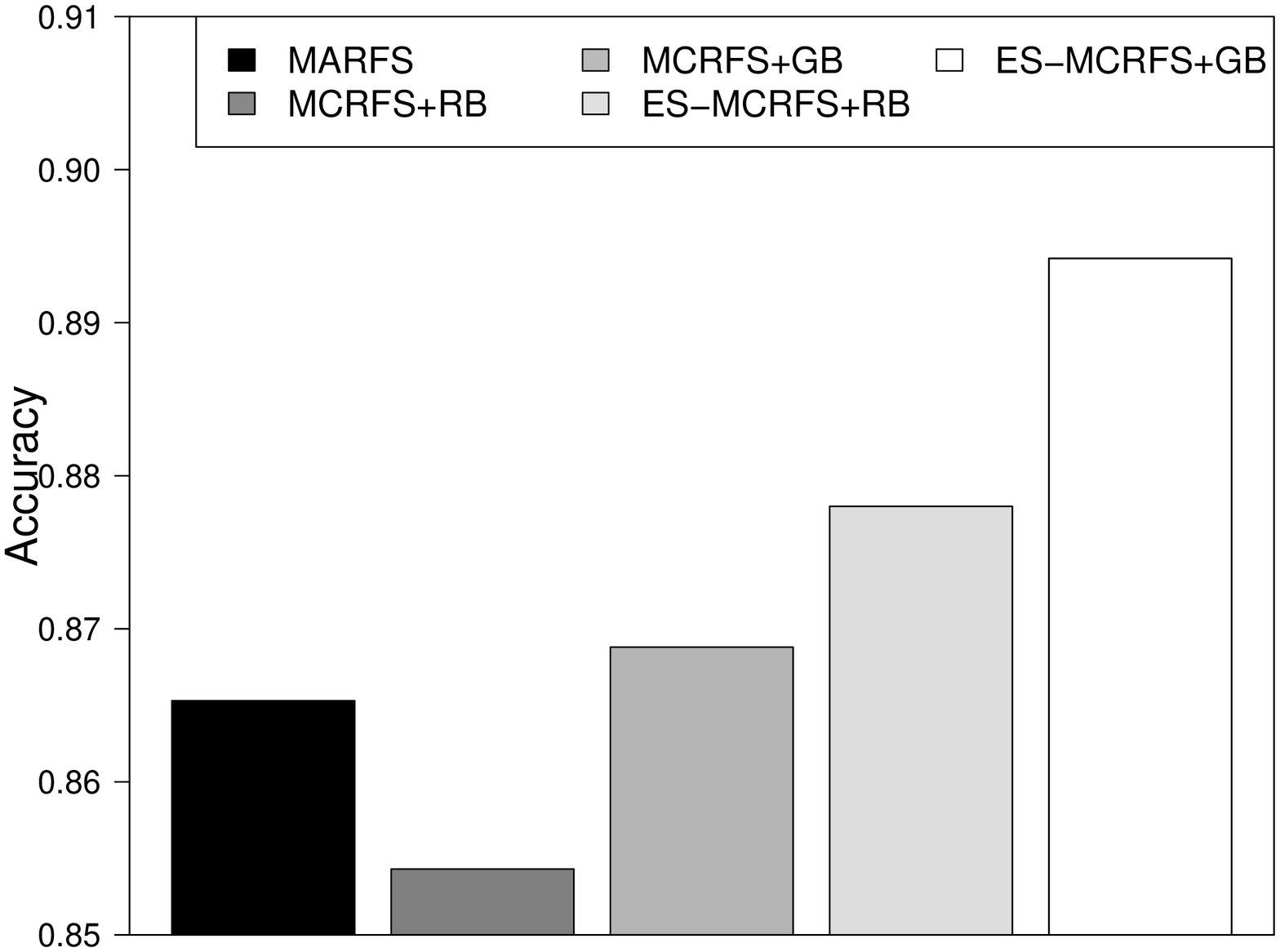}
}
\hspace{-4mm}
\subfigure[Spam]{ 
\includegraphics[width=4cm]{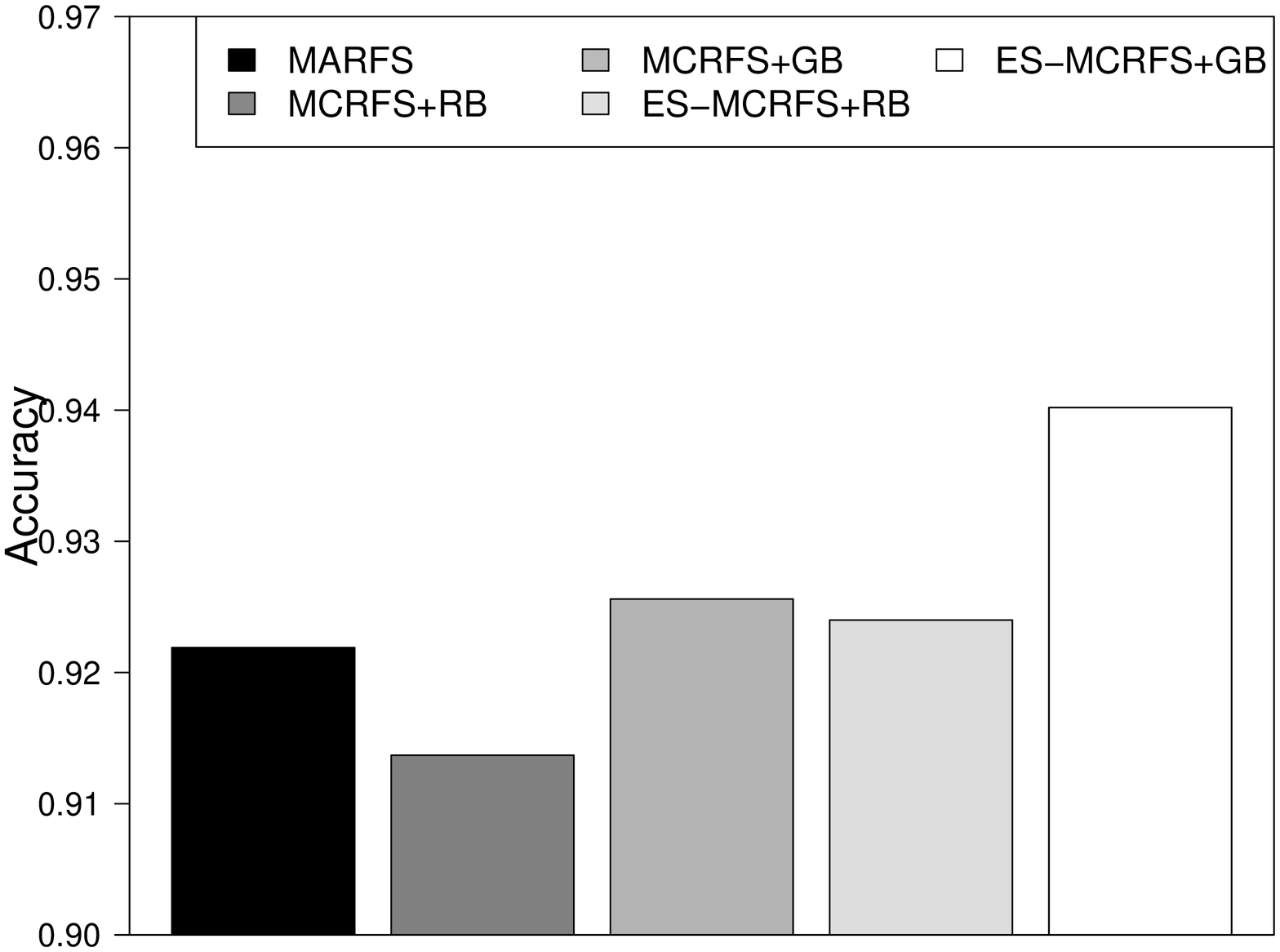}
}
\vspace{-4mm}
\subfigure[ICB]{
\includegraphics[width=4.1cm]{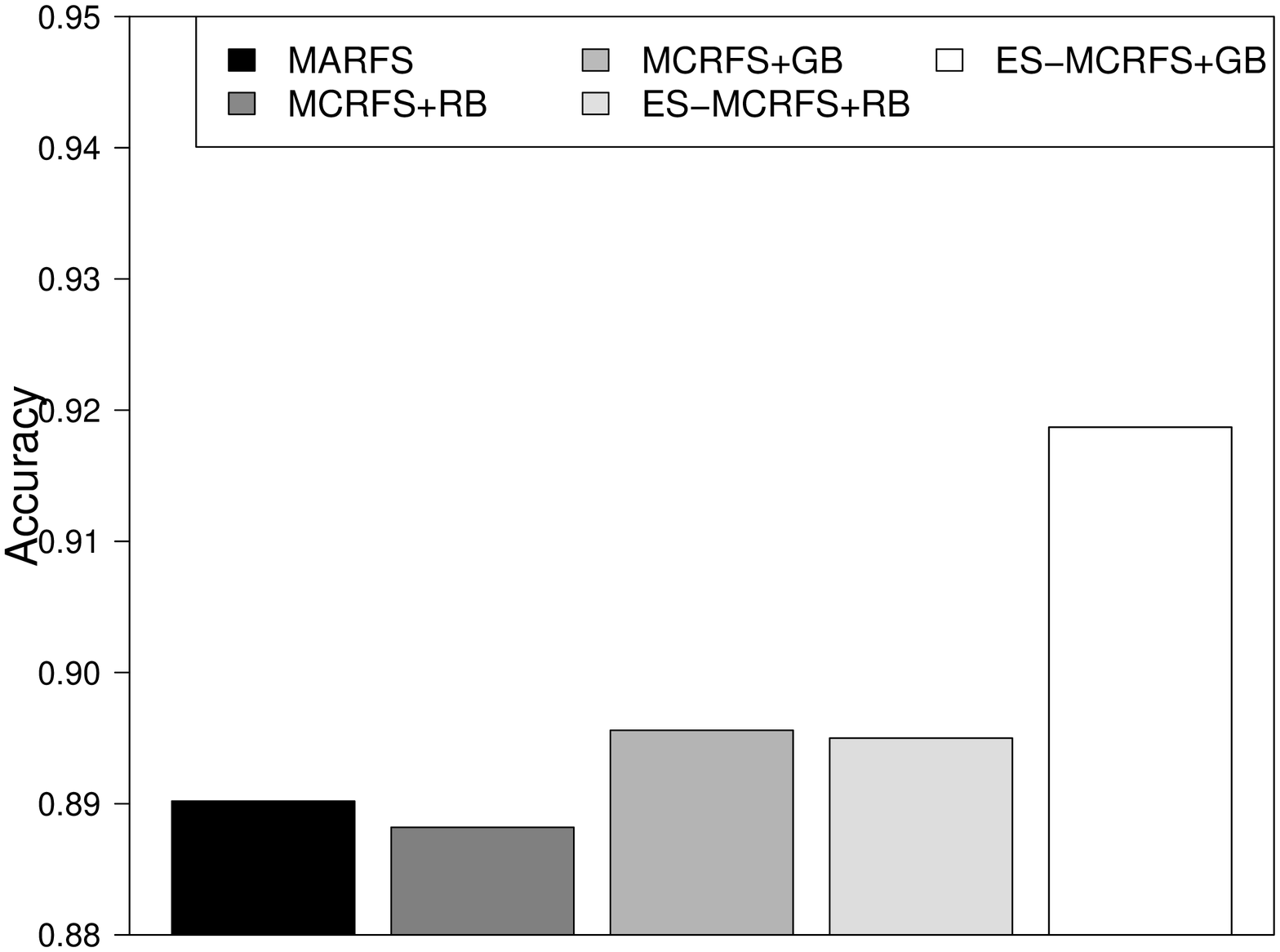}
}
\hspace{-4mm}
\subfigure[Arrhy]{ 
\includegraphics[width=4.1cm]{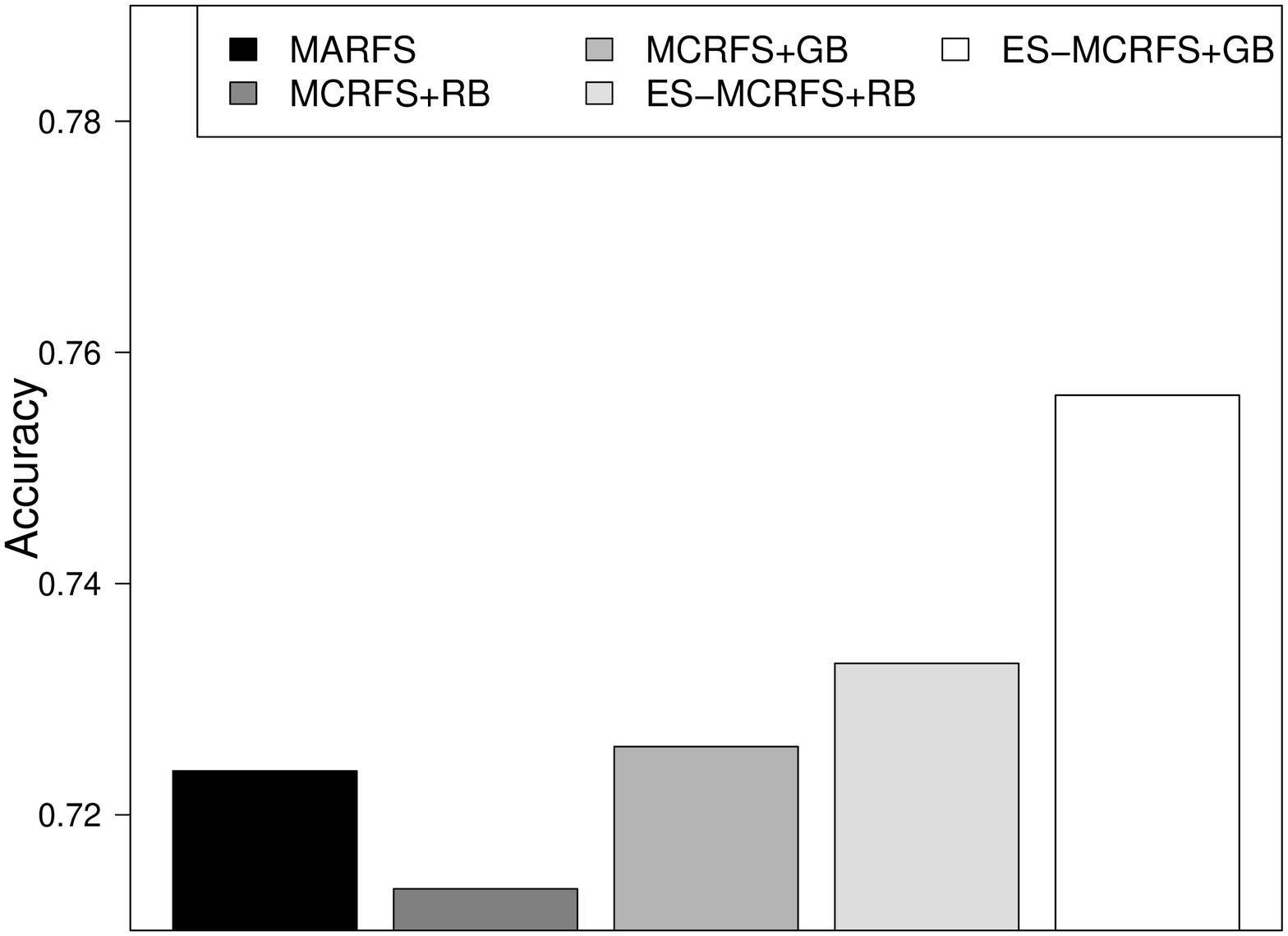}
}
\caption{Predictive accuracy on different training strategies. RB for random behavior policy and GB for $\epsilon$-greedy behavior policy.}
\label{behaviro_policy}
\end{figure}

We study the difference between random behavior policy and the $\epsilon$-greedy policy presented in Equation \ref{pro_b}. We combine the two policies with MCRFS and ES-MCRFS respectively. Figure \ref{behaviro_policy} shows that the $\epsilon$-greedy policy outperforms the random behavior policy on all datasets.

\subsection{Computational Burden of Traverse Strategy}
We compare the computational burden of the MCRFS which uses single-agent and the traverse strategy to substitute the multi-agent strategy in the MARFS. Table \ref{CPU_Mem_Occ} shows that the CPU and memory cost when implementing the two methods. Our method MCRFS requires less computational resources than the multi-agent MARFS.

\begin{table}[h]
\centering
\scriptsize
\caption{ \small CPU and memory (in MB) occupation.}
\label{CPU_Mem_Occ}

\begin{tabular}{|c|c|c|c|c|c|c|c|c|}
\hline
\multirow{2}{*}{} & \multicolumn{2}{c|}{FC} & \multicolumn{2}{c|}{Spam} & \multicolumn{2}{c|}{ICB} & \multicolumn{2}{c|}{Arrhy} \\ \cline{2-9} 
                  & CPU        & Mem        & CPU         & Mem         & CPU         & Mem        & CPU          & Mem         \\ \hline
MARFS             & 72\%       & 1531       & 75\%        & 1502        & 86\%        & 1797       & 97\%         & 4759        \\ \hline
MCRFS             & 57\%       & 1429       & 54\%        & 1395        & 59\%        & 1438       & 55\%         & 1520        \\ \hline
\end{tabular}
\end{table}

\subsection{Decision History Based Traverse Strategy}
We study the decision history based traverse strategy by comparing its performance with the vanilla traverse strategy on ES-MCRFS. Table \ref{Deci_His} shows that the decision history can significantly improve performance of the traverse strategy.

\begin{table}[h]
 \setlength{\arraycolsep}{1.5pt}
\scriptsize
\centering
\caption{\small Traverse strategy ablation. DH for decision history.}
\label{Deci_His}

\begin{tabular}{|c|c|c|c|c|c|c|c|c|}
\hline
\multirow{2}{*}{}           & \multicolumn{2}{c|}{FC}                               & \multicolumn{2}{c|}{Spam}                             & \multicolumn{2}{c|}{ICB}                              & \multicolumn{2}{c|}{Arrhy}                            \\ \cline{2-9} 
                            & Acc                       & F1                        & Acc                       & F1                        & Acc                       & F1                        & Acc                       & F1                        \\ \hline
\multicolumn{1}{|l|}{No DH} & \multicolumn{1}{l|}{0.75} & \multicolumn{1}{l|}{0.82} & \multicolumn{1}{l|}{0.83} & \multicolumn{1}{l|}{0.79} & \multicolumn{1}{l|}{0.75} & \multicolumn{1}{l|}{0.81} & \multicolumn{1}{l|}{0.59} & \multicolumn{1}{l|}{0.56} \\ \hline
With DH                     & 0.89                      & 0.88                      & 0.94                      & 0.91                      & 0.92                      & 0.88                      & 0.76                      & 0.74                      \\ \hline
\end{tabular}
\end{table}

\begin{table*}[t!]
\centering
\caption{Performance with different utility function}
\label{utility_func}
\setlength{\tabcolsep}{4mm}{
\begin{tabular}{|l|l|l|l|l|l|l|l|l|l|}
\hline
\multicolumn{2}{|l|}{\multirow{2}{*}{}}    & \multicolumn{2}{c|}{FC}                            & \multicolumn{2}{c|}{Spam}                          & \multicolumn{2}{c|}{ICB}                           & \multicolumn{2}{c|}{Arrhy}                         \\ \cline{3-10} 
\multicolumn{2}{|l|}{}                     & \multicolumn{1}{c|}{Acc} & \multicolumn{1}{c|}{F1} & \multicolumn{1}{c|}{Acc} & \multicolumn{1}{c|}{F1} & \multicolumn{1}{c|}{Acc} & \multicolumn{1}{c|}{F1} & \multicolumn{1}{c|}{Acc} & \multicolumn{1}{c|}{F1} \\ \hline
\multirow{3}{*}{\begin{sideways}{{Utility}}\end{sideways}} & $Rd$    & 0.8689                   & 0.8433                  & 0.9250                   & 0.8749                  & 0.8997                   & 0.8788                  & 0.7340                   & 0.6955                  \\ \cline{2-10} 
                                   & $Rv$    & 0.8703                   & 0.8507                  & 0.9317                   & 0.8831                  & 0.9001                   & 0.8793                  & 0.7393                   & 0.7143                  \\ \cline{2-10} 
                                   & $Rv-Rd$ & \textbf{0.8842}          & \textbf{0.8650}         & \textbf{0.9402}          & \textbf{0.8949}         & \textbf{0.9117}          & \textbf{0.8903}         & \textbf{0.7492}          & \textbf{0.7258}         \\ \hline
\end{tabular}
}
\end{table*}

\vspace{-0.1cm}

\subsection{Training Efficiency of reward-level interactive strategy}

\begin{figure}[h]
\centering
\subfigure[FC]{
\includegraphics[width=4.1cm]{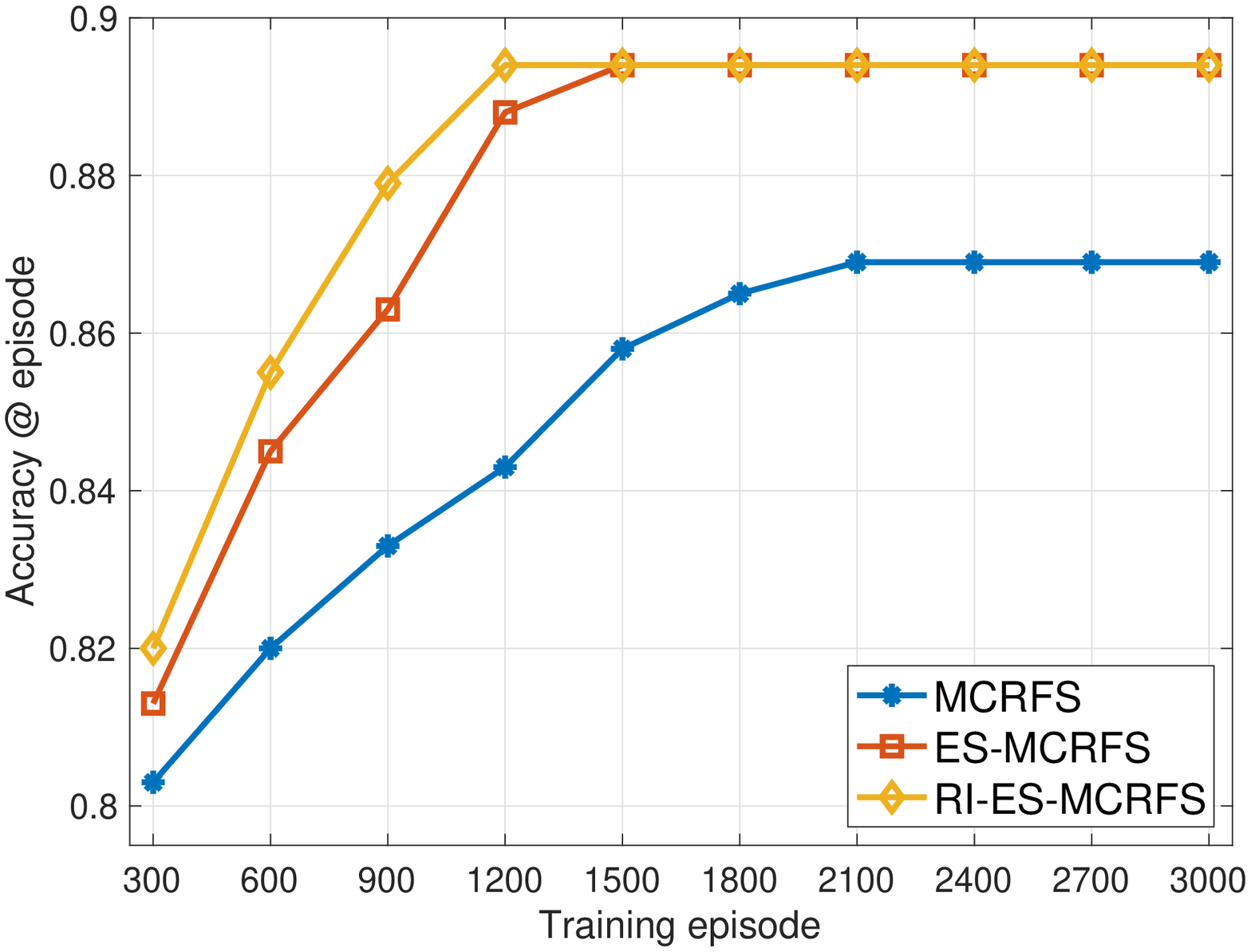}
}
\hspace{-5mm}
\subfigure[Spam]{ 
\includegraphics[width=4.1cm]{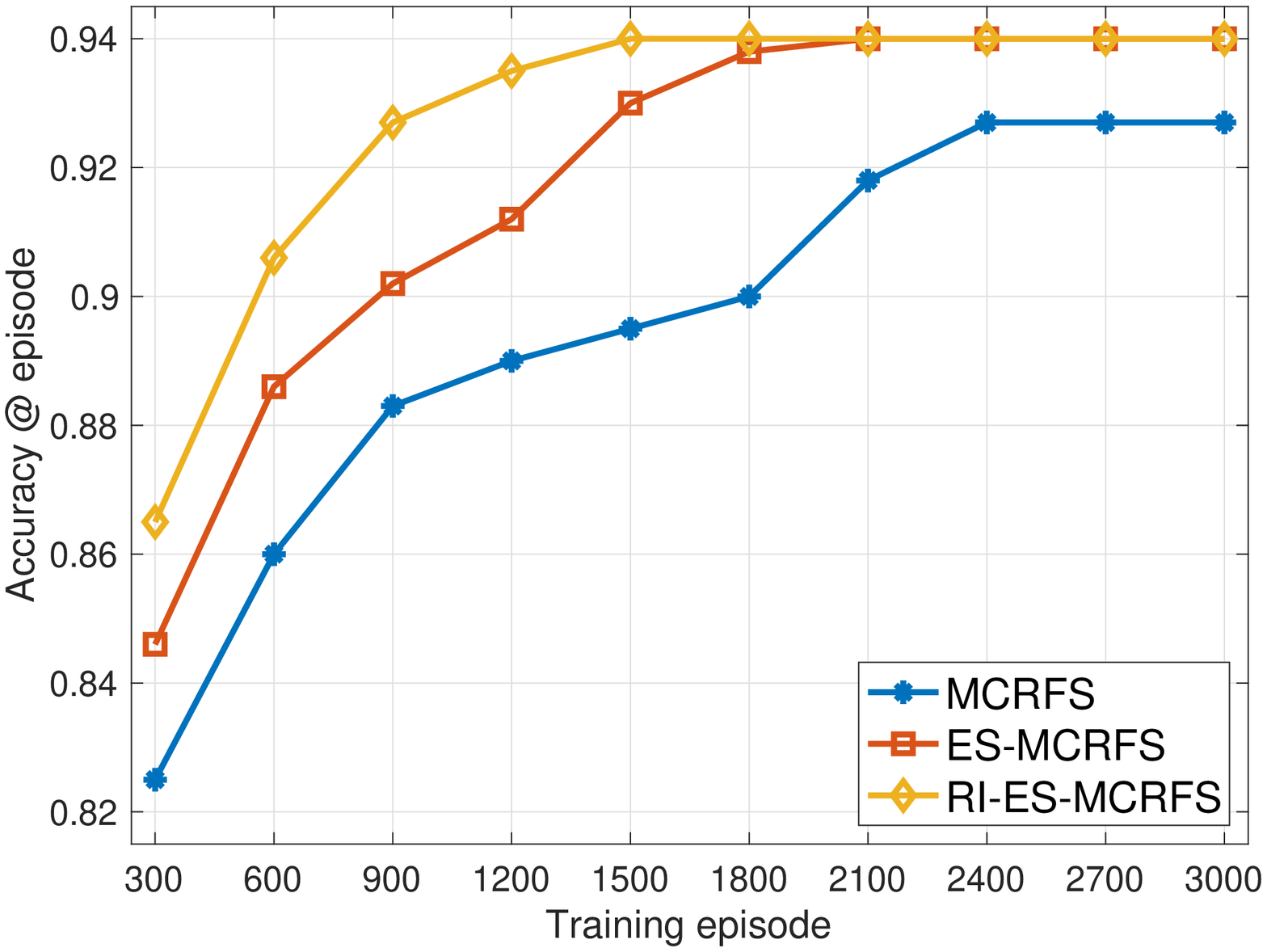}
}
\vspace{-4mm}
\subfigure[ICB]{
\includegraphics[width=4.1cm]{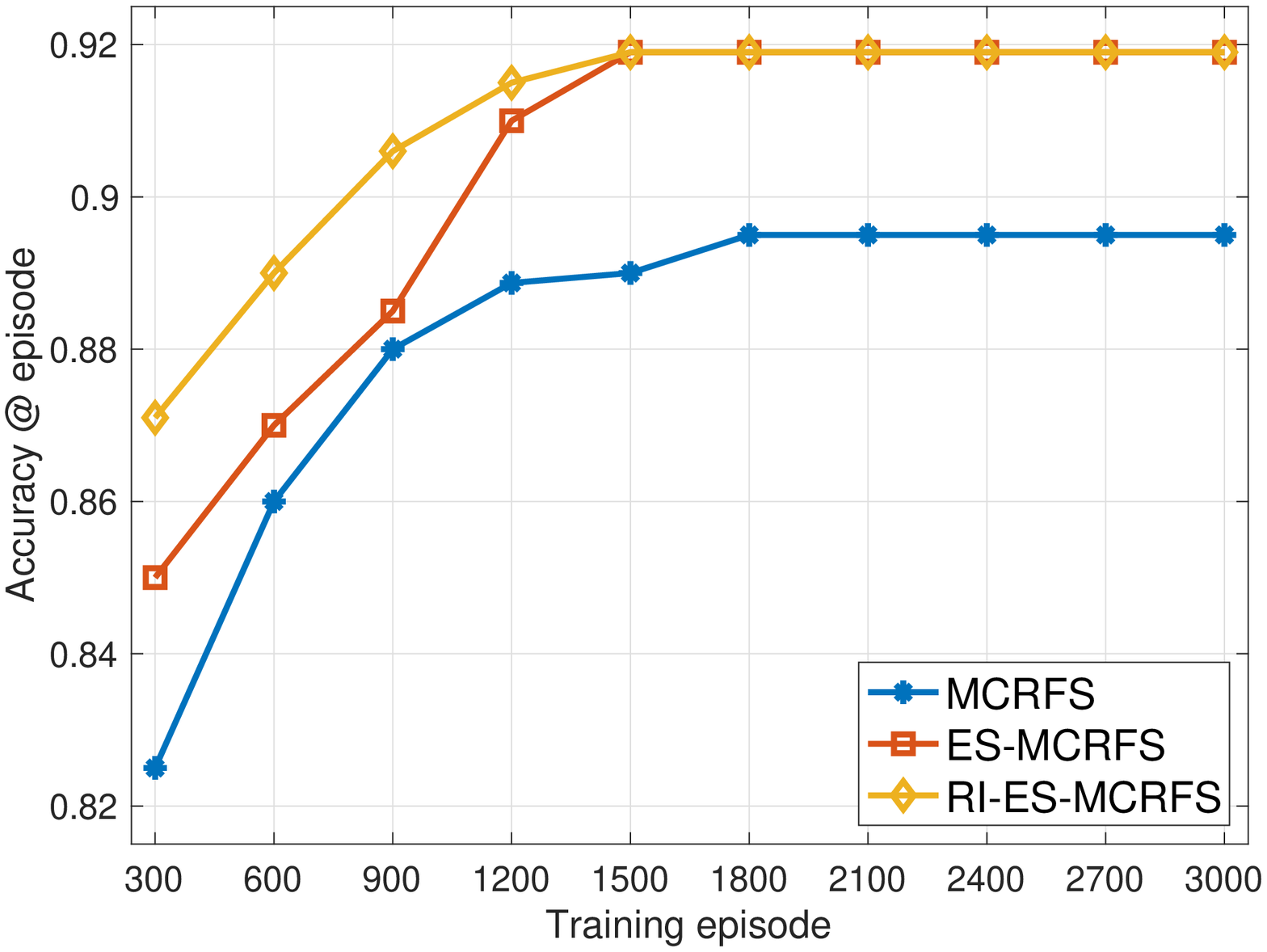}
}
\hspace{-5mm}
\subfigure[Arrhy]{ 
\includegraphics[width=4.1cm]{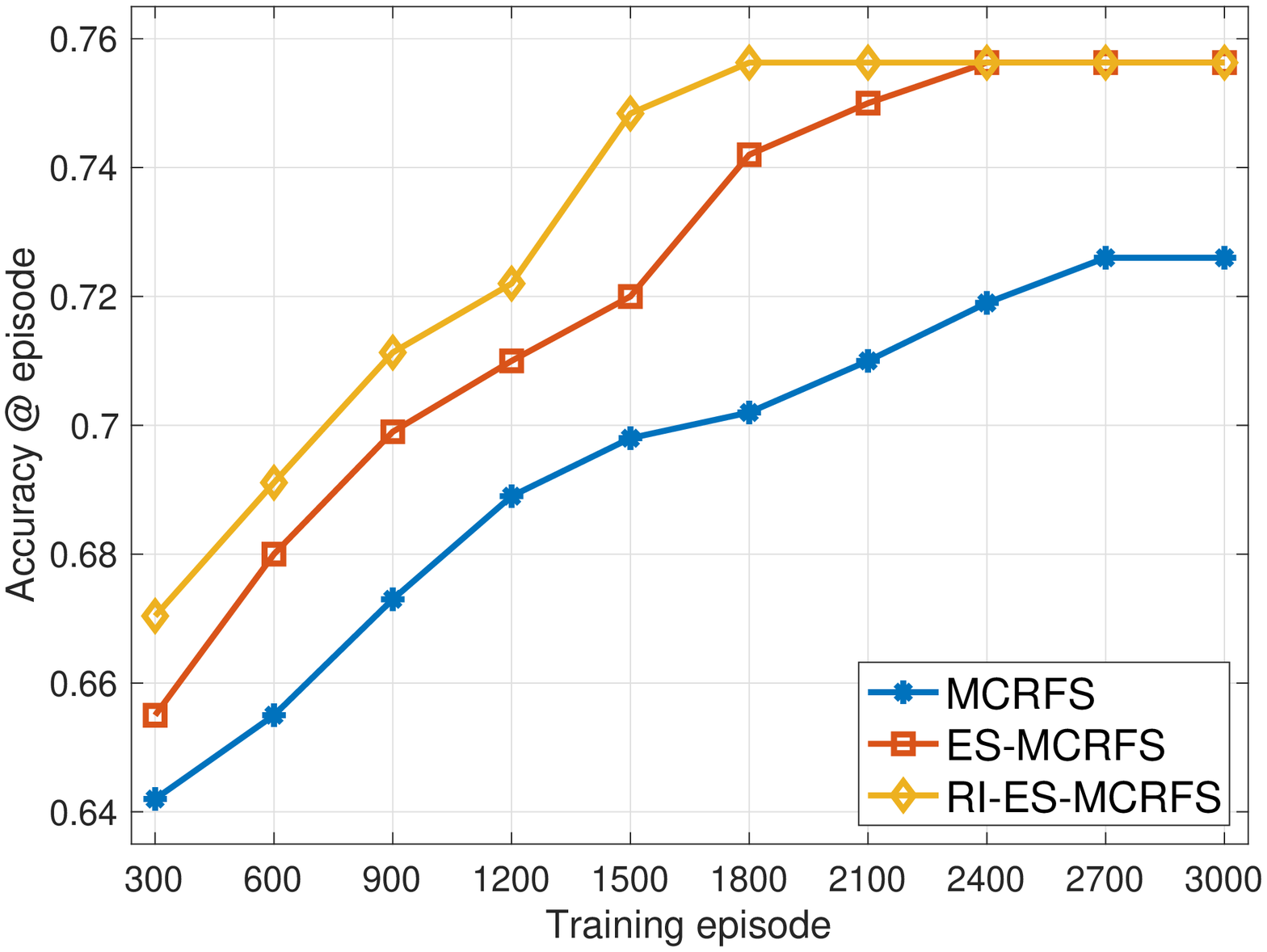}
}
\caption{Predictive accuracy on training step.}
\label{reward_interactive}
\end{figure}

We compare the predictive accuracy with different numbers of training episodes to study the training efficiency of the reward-level interactive (RI) strategy. Figure \ref{reward_interactive} shows that with RI, the Monte Carlo reinforced feature selection can achieve convergence more quickly. However, as the ES-MCRFS already achieves good performance, the  RI can not improve its final performance.

\vspace{-0.1cm}
\subsection{Study of the Utility Function}
We define the utility function $\mathcal{U}$ as the combination of relevance ($Rv$) function and redundancy ($Rd$) function. Here we study the impact of the two components for the utility function. Table \ref{utility_func} shows that when we use $Rv$ independently as the utility function, its performance is better than the $Rd$. This is because $Rv$ evaluates the relationship between features and the label, which is directly related to the classification task, while $Rd$ evaluates the relationship among features, which is an indirect evaluation to the classification task. The combination of the two functions ($Rv-Rd$) as the utility function significantly outperforms each of the independent functions, revealing the $Rd$ and $Rv$ coordinate and make up each other's shortage.

\section{Related Work}\label{related_work}

\noindent \textbf{Efficient Sampling in Reinforcement Learning.} 
Reinforcement learning is a trial-and-error based method, which requires high-quality samples to train its policy. It is always a hot topic to pursue efficient sampling for reinforcement learning. One research direction is to generate training samples with high quality based on the importance sampling technology, 
such as rejection control \cite{liu1998rejection} and marginalized importance sampling \cite{xie2019towards}. 
These methods basically control the sampling process based on the importance sampling weight. Another research direction is to sample diversified sample from different policy parameters. The diversity partially contributes to the exploration and thus have better performance on some specific tasks \cite{fortunato2017noisy}.
However, these methods suffer from slow convergence and no theoretical guarantee \cite{yu2018towards}. Besides, there are other attempts to develop sample efficient reinforcement learning, such as curiosity-driven exploration and hybrid optimization \cite{raginsky2017non,wang2020defending}.

\noindent \textbf{Feature Selection.} 
Feature selection can be categorized into three types, i.e., filter methods, wrapper methods and embedded methods \cite{liu2021automated,zhao2020simplifying}.
Filter methods rank features only by relevance scores and only top-ranking features are selected. The representative filter methods is the  univariate feature selection \cite{forman2003extensive} 
The representative wrapper methods are branch and bound algorithms \cite{narendra1977branch,kohavi1997wrappers}.  
Wrapper methods are supposed to achieve better performance than filter methods since they search on the whole feature subset space. Evolutionary algorithms \cite{yang1998feature,kim2000feature} low down the computational cost but could only promise local optimum results.
Embedded methods combine feature selection with predictors more closely than wrapper methods. The most widely used embedded methods are LASSO \cite{tibshirani1996regression} and decision tree \cite{sugumaran2007feature}. 

\noindent \textbf{Interactive Reinforcement Learning.} 
Interactive reinforcement learning (IRL) is proposed to accelerate the learning process of reinforcement learning. Early work on the IRL topic can be found in \cite{lin1991programming}, where the authors presents a general approach to making robots which can improve their performance from experiences as well as from being taught. Unlike the imitation learning which intends to learn from an expert other than the environment \cite{schaal1999imitation,ho2016generative}, IRL sticks to learning from the environment and the advisor is only an advice-provider in its apprenticeship \cite{knox2013teaching,wang2021reinforced}. 
As the task for the advisor is to help the agent pass its apprenticeship, the advisor has to identify which states belong to the apprenticeship. In \cite{torrey2013teaching}, the authors study the advising state selection and propose four advising strategies, i.e., early advising, importance advising, mistake correcting and predictive advising. 


\section{Conclusion Remarks}

\noindent \textbf{Summary.}
In this paper, we study the problem of improving the training efficiency of reinforced feature selection (RFS). We propose a traverse strategy to simplify the multi-agent formulation of the RFS to a single-agent framework, an implementation of Monte Carlo method under the framework, and two strategies to improve the efficiency of the framework. 

\noindent\textbf{Theoretical Implications.}
The single-agent formulation reduces the requirement of computational resources, the early stopping strategy improves the training efficiency, the decision history based traversing strategy diversify the training process, and the interactive reinforcement learning accelerates the training process without changing the optimal policy.

\noindent \textbf{Practical Implications.}
Experiments show that the Monte Carlo method with the traverse strategy can significantly reduce the hardware occupation in practice, the decision history based traverse strategy can improve performance of the traverse strategy, the interactive reinforcement learning can improve the training of the framework.

\noindent \textbf{Limitations and Future Work.}
Our method can be further improved from the following aspects: 1) The framework can be adapted into a parallel framework, where more than one (but much smaller than the feature number) agents work together to finish the traverse; 2) Besides reward level, the interactive reinforcement learning can obtain advice from action level and sampling level.
3) The framework can be implemented on any other reinforcement learning frameworks, e.g., deep Q-network, actor critic and proximal policy optimization (PPO).

\section*{Acknowledgements}
This research was partially supported by the National Science Foundation (NSF) via the grant numbers: 2008837,  2007210, 1755946, I2040950 and 2006889.
\bibliographystyle{IEEEtran}
\bibliography{ref}

\end{document}